\documentclass[12pt]{article}
\usepackage{amsmath}
\usepackage{amssymb}
\usepackage{amsthm}
\usepackage[pdftex]{graphicx}
\usepackage{psfrag,epsf}
\usepackage{enumerate}
\usepackage{natbib}
\usepackage{subcaption}
\usepackage[colorlinks,citecolor=blue,urlcolor=blue]{hyperref}
\allowdisplaybreaks
\usepackage[boxruled,scleft]{algorithm2e}

\usepackage[top=1in, bottom=1in, left=1in, right=1in]{geometry}

\newcommand{\argmax}{\operatornamewithlimits{arg \,max}}
\providecommand{\tightlist}{%
  \setlength{\itemsep}{0pt}\setlength{\parskip}{0pt}}
\RestyleAlgo{boxruled}

\numberwithin{equation}{section}

\newtheorem{thm}{Theorem}
\newtheorem{lem}{Lemma}
\newtheorem{prop}{Proposition}

\bibpunct{(}{)}{;}{a}{}{,}
\begin{document}

\begin{center}
{\large \bf Higher-Order Spectral Clustering under Superimposed Stochastic Block Models
}

\vspace{0.2in}

Subhadeep Paul, The Ohio State University\footnote{Emails: Paul - {\it paul.963@osu.edu}, Milenkovic -  {\it milenkov@illinois.edu}, Chen - {\it yuguo@illinois.edu}}

\vspace{0.05in}

Olgica Milenkovic, University of Illinois at Urbana-Champaign

\vspace{0.05in}

Yuguo Chen, University of Illinois at Urbana-Champaign

\end{center}

\begin{abstract}
Higher-order motif structures and multi-vertex interactions are becoming increasingly important in studies that aim to improve our understanding of functionalities and evolution patterns of networks. To elucidate the role of higher-order structures in community detection problems over complex networks, we introduce the notion of a Superimposed Stochastic Block Model (SupSBM). 
 The model is based on a random graph framework in which certain higher-order structures or subgraphs are generated through an independent hyperedge generation process, and are then replaced with graphs that are superimposed with directed or undirected edges generated by an inhomogeneous random graph model. Consequently, the model introduces controlled dependencies between edges which allows for capturing more realistic network phenomena, namely strong local clustering in a sparse network, short average path length, and community structure. We proceed to rigorously analyze the performance of a number of recently proposed higher-order spectral clustering methods on the SupSBM. In particular, we prove non-asymptotic upper bounds on the misclustering error of spectral community detection for a SupSBM setting in which triangles or 3-uniform hyperedges are superimposed with undirected edges. As part of our analysis, we also derive new bounds on the misclustering error of  higher-order spectral clustering methods for the standard SBM and the 3-uniform hypergraph SBM. Furthermore, for a non-uniform hypergraph SBM model in which one directly observes both edges and 3-uniform hyperedges, we obtain a criterion that describes when to perform spectral clustering based on edges and when on hyperedges, based on a function of hyperedge density and observation quality.
\end{abstract}


\section{Introduction}

Network data science has traditionally focused on studies capturing two-way interactions or connections between pairs of vertices or agents in networks. In this context, the problems of interest have been to identify heterogeneous and power law vertex degree distributions (e.g., determine if the networks are scale-free) as well as dense subgraphs and cliques, and efficiently detect and isolate community structures~\citep{newman2003structure,barabasi1999emergence,watts1998collective}. 

It has by now become apparent that many aspects of relational organization, functionality and the evolving structure of a complex network can only be understood through higher-order subgraph (motif) interactions involving more than two vertices~\citep{milo2002network,shen2002network,mangan2003structure,honey2007network,alon2007network,porter2009communities,benson2016higher,yaverouglu2014revealing}. Certain subgraphs in networks function as fundamental units of control and regulation of network communities and dynamics: for example, network motifs are crucial regulators in brain networks~\citep{sporns2004motifs,park2013structural,battiston2017multilayer}, transcriptional regulatory networks~\citep{mangan2003structure}, food webs \citep{paulau2015motif,li2017inhomogoenous}, social networks~\citep{girvan2002community,snijders2001statistical} and air traffic networks~\citep{rosvall2014memory,benson2016higher}. Traditionally, statistical and algorithmic work on network motifs has been concerned with discovering and counting the frequency of over-expressed subgraphs (which are usually determined in comparison with some statistical null model) in various real world networks~\citep{alon2007network,klusowski2018counting}.  
Indeed, frequency distributions or spectra of motifs have been shown to provide useful information about the regulatory and dynamic organization of networks obtained from disparate sources. Network motifs have also recently been used to perform learning tasks such as community detection~\citep{benson2016higher,li2017inhomogoenous,tsourakakis2017scalable}. A parallel line of work has focused on identifying communities in hypergraphs and was reported in~\citet{zhou2006learning,angelini2015spectral,kim2017community,ghoshdastidar2017consistency,chien2018community}.

Unfortunately, existing random graph models with community structures based on Erd\"os-R\'enyi random graphs \citep{er60}, such as the Stochastic Block Models~\citep{hll83,sn97,bc09,cwa12,rqf12, cdp11,rcy11,qr13,j15,lei2015consistency,decelle2011asymptotic,hajek2016achieving,abbe2015community,gao2017achieving}, their degree-corrected versions~\citep{kn11,zlz12}, and other extensions fail to produce graphs with strong local clustering, i.e., with over-abundant triangles and other relevant higher-order structures. To investigate community structures in terms of particular subgraphs and determine under which conditions they can be recovered or detected, one needs to consider more versatile community structure models.

To address the aforementioned problem, a number of more realistic network models with some of the desired motif structures have been proposed in the literature; however, most such models are not mathematically tractable in general or in the context of community detection due to dependencies among the edges~\citep{bollobas2011sparse}. Notable exceptions include the mathematically tractable random graph model with local clustering and dependences among edges proposed in~\cite{bollobas2011sparse}. There, the authors constructed random graphs by superimposing small subgraphs and edges, thereby introducing dependencies among subsets of vertices. More specifically, they constructed an inhomogeneous random hypergraph with conditionally independent
hyperedges, and then replaced each hyperedge by a complete graph over the same set of vertices. A similar model, termed the Subgraph Generation Model (SUGM), was proposed in~\cite{chandrasekhar2014tractable,chandrasekhar2016network}. 

More recently,~\citet{hajek2018recovering} analyzed a variation of the preferential attachment model with community structure and proposed a message passing algorithm to recover the communities. In parallel, a geometric block model that uses Euclidean latent space geometric graphs instead of the usual Erdo\"s-Re\'nyi graphs for the mixture components was introduced in~\citet{galhotra2017geometric,galhotra2018connectivity}. Although all these modes capture some aspects of real-life networks and introduce controlled dependencies among the edges in the graphs, they fail to provide a general approach for combining dependent motif structures and analytical techniques that highlight if communities should be identified though pairwise or higher-order interactions.  

Our contributions are two-fold.  
First, we propose a new Superimposed Stochastic Block Model (SupSBM), a random graph model for networks with community structure obtained  by generalizing the framework of~\cite{chandrasekhar2014tractable} and~\cite{bollobas2011sparse} to account for communities akin to the classical SBM. SupSBM captures the most relevant aspects of higher-order organization of the datasets under consideration, e.g., it incorporates triangles and other motifs, but couples them through edges that may be viewed as noise in the motif-based graphs. The community structure of interest maybe present either at a higher-order structural level only or both at the level of higher-order structures and edges. Drawing parallels with the classical SBM which is a mixture of Erd\"os-R\'enyi graphs, SupSBM may be viewed as a mixture of superimposed inhomogeneous random graphs generated according to process described in~\cite{chandrasekhar2014tractable} and~\cite{bollobas2011sparse}. Second, we derive theoretical performance guarantees for higher-order spectral clustering methods~\citep{benson2016higher,tsourakakis2017scalable} applied to SupSBM. The main difference between our analysis and previous lines of work on spectral algorithms for the SBM \citep{rcy11,lei2015consistency,  gao2017achieving, chin2015stochastic, vu13}, and hypergraph SBM \citep{ghoshdastidar2017consistency, kim2017community,chien2018community} is that the elements of the analogues of adjacency matrices in our analysis are dependent. We derive several non-asymptotic upper bounds of the spectral norms of such generalized adjacency matrices, and these results are of independent interest in other areas of network analysis. For this purpose, we express the spectral norms as sums of polynomial functions of independent random variables. The terms in the sums are dependent, however, any given term is dependent only on a small fraction of other terms. We exploit this behavior to carefully control the effects of such dependence on the functions of interest. We use recent results on polynomial functions of independent random variables~\citep{boucheron2013concentration,kim2000concentration,janson2004deletion}, typical bounded differences inequalities~\citep{warnke2016method} and Chernoff style concentration inequalities under limited dependence~\citep{warnke2017upper} to complete our analysis. In addition, we derive a number of corollaries implying performance guarantees for higher-order spectral clustering under the classical stochastic block model and the hypergraph stochastic block model. The analysis of the non-uniform hypergraph SBM also reveals interesting results regarding the benefit of using ordinary versus higher order spectral clustering methods on random hypergraphs.

The remainder of the article is organized as follows. Section~\ref{sec:super} defines superimposed random graph models and then develops the Superimposed Stochastic Block Model (SupSBM). Section~\ref{sec:analysis} presents a non-asymptotic analysis of the misclustering rate of higher-order spectral clustering under the SupSBM. Some real world network examples are discussed in Section~\ref{sec:data}. The Appendix contains proofs of all the theorems and many auxiliary lemmas used in the derivations.

\section{Superimposed random graph and block models} \label{sec:super}

We start our analysis by defining what we refer to as an \emph{inhomogeneous superimposed random graph model,} which is based on the random graph models described in~\citet{bollobas2011sparse,chandrasekhar2014tractable}. We then proceed to introduce a natural extension of the stochastic block model in which the community components are superimposed random graphs. Our main focus is on models that superimpose edges and triangles, as these are prevalent motifs in real social and biological networks \citep{alon2007network,benson2016higher,li2017inhomogoenous,laniado2016gender}.
However, as discussed in subsequent sections, the superimposed SBM can be easily extended to include other superimposed graph structures. 

Formally, the proposed random graph model, denoted by $G_s(n,P^e,\mathbb{P}^t)$, is a superimposition of a classical dyadic (edge-based) random graph $G_e(n,P^e)$ and a triadic (triangle-based) random graph $G_t(n, \mathbb{P}^t)$. In this setting, $n$ denotes the number of vertices in the graph, $P^e$ denotes an $n \times n$ matrix whose $(i,j)$th entry equals the probability of an edge in $G_e$ between the vertices $i$ and $j$, and $\mathbb{P}^t$ denotes a $3$-way ($3$rd order) $n \times n \times n$ tensor whose $(i,j,k)$th element equals the probability of a triangle involving the vertices $(i,j,k)$ in $G_t$. 

A random graph from the model $G_s(n,P^e,\mathbb{P}^t)$ is generated as follows. One starts with $n$ unconnected vertices. The $G_t(n, \mathbb{P}^t)$ graph is generated by creating triangles ($3-$hyperedges) for each of the $\dbinom{n}{3}$ $3$-tuples of vertices $(i,j,k)$ according to the outcome of \emph{independent} Bernoulli random variables $T_{ijk}$ with parameter $p^t_{ijk}=(\mathbb{P}^t)_{ijk}$. The hyperedges are consequently viewed as triangles, which results in a loss of their generative identity. Note that this process may lead to multi-edges between pairs of vertices $i$ and $j$ if these are involved in more than one triangle. The multi-edges in the graph $G_t$ are collapsed into single edges. All pairs of vertices $(i,j)$ remain within all their constituent triangles as before the merging procedure. Next, the graph $G_e(n,P^e)$ is generated by placing edges between the $\dbinom{n}{2}$ pairs of vertices $(i,j)$ according to the outcomes of independent Bernoulli random variables $E_{ij}$ with parameter $p^e_{ij}=(P^e)_{ij}$. Note this is simply the usual inhomogeneous random graph model~\citep{bollobas2007phase} that may be viewed as a generalization of the Erd\"os-R\'enyi model in which the probabilities of individual edges are allowed to be nonuniform. 
The two independently generated graphs are then superimposed to arrive at $G_s(n,P^e,\mathbb{P}^t)$. 

\begin{figure}[h]
\centering
\begin{subfigure}{0.16\textwidth}
\centering
\includegraphics[width=\linewidth]{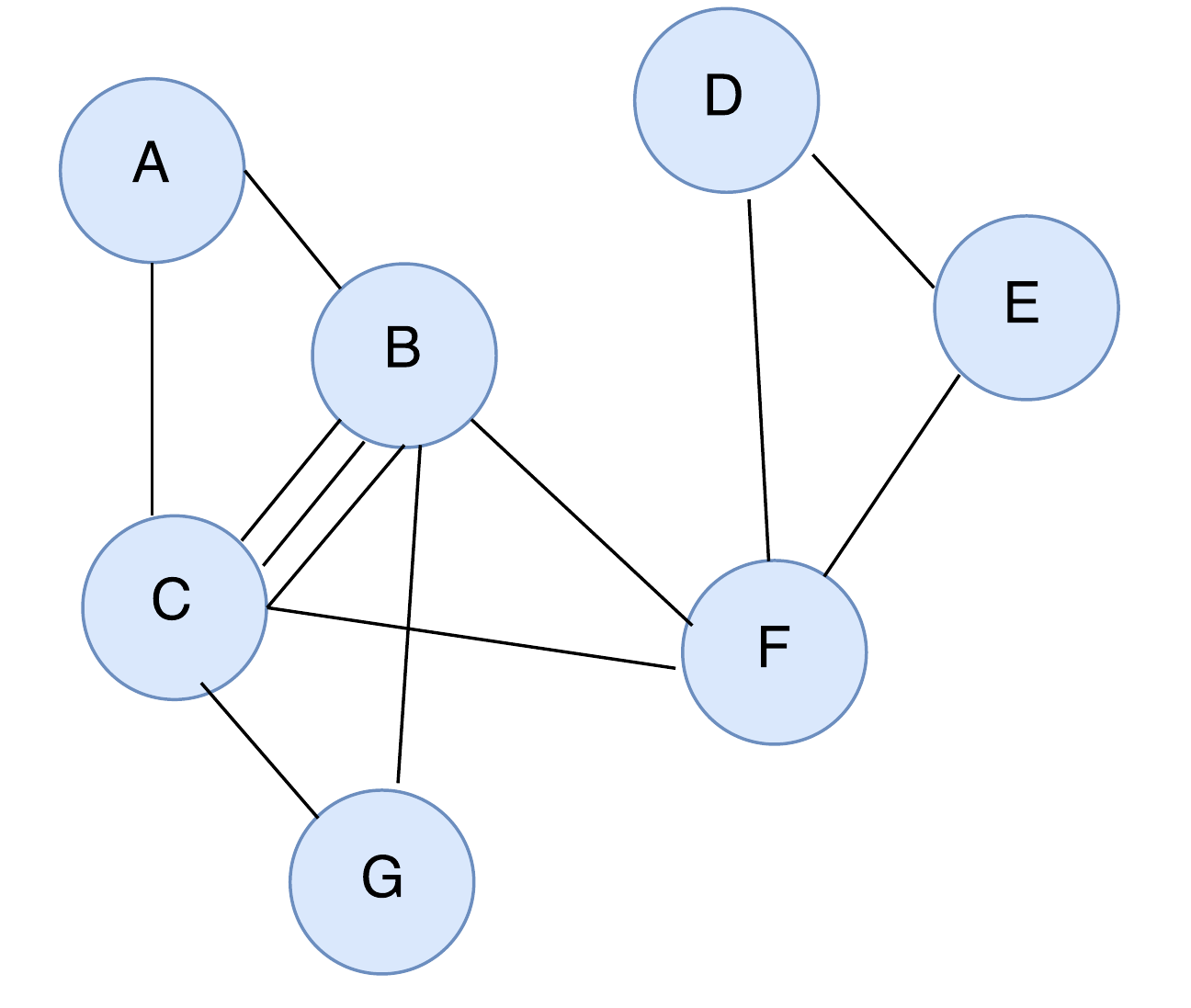}
 \end{subfigure}%
 \begin{subfigure}{0.05\textwidth}
\centering
\includegraphics[width=\linewidth]{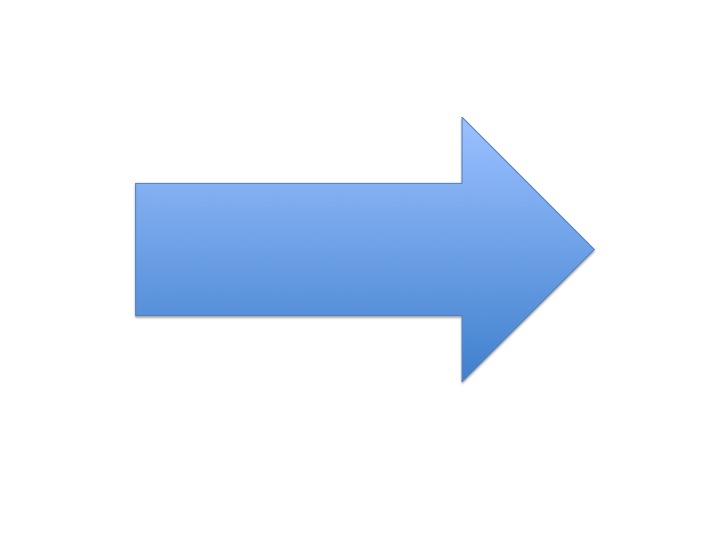}
 \end{subfigure}%
 \begin{subfigure}{0.16\textwidth}
\centering
\includegraphics[width=\linewidth]{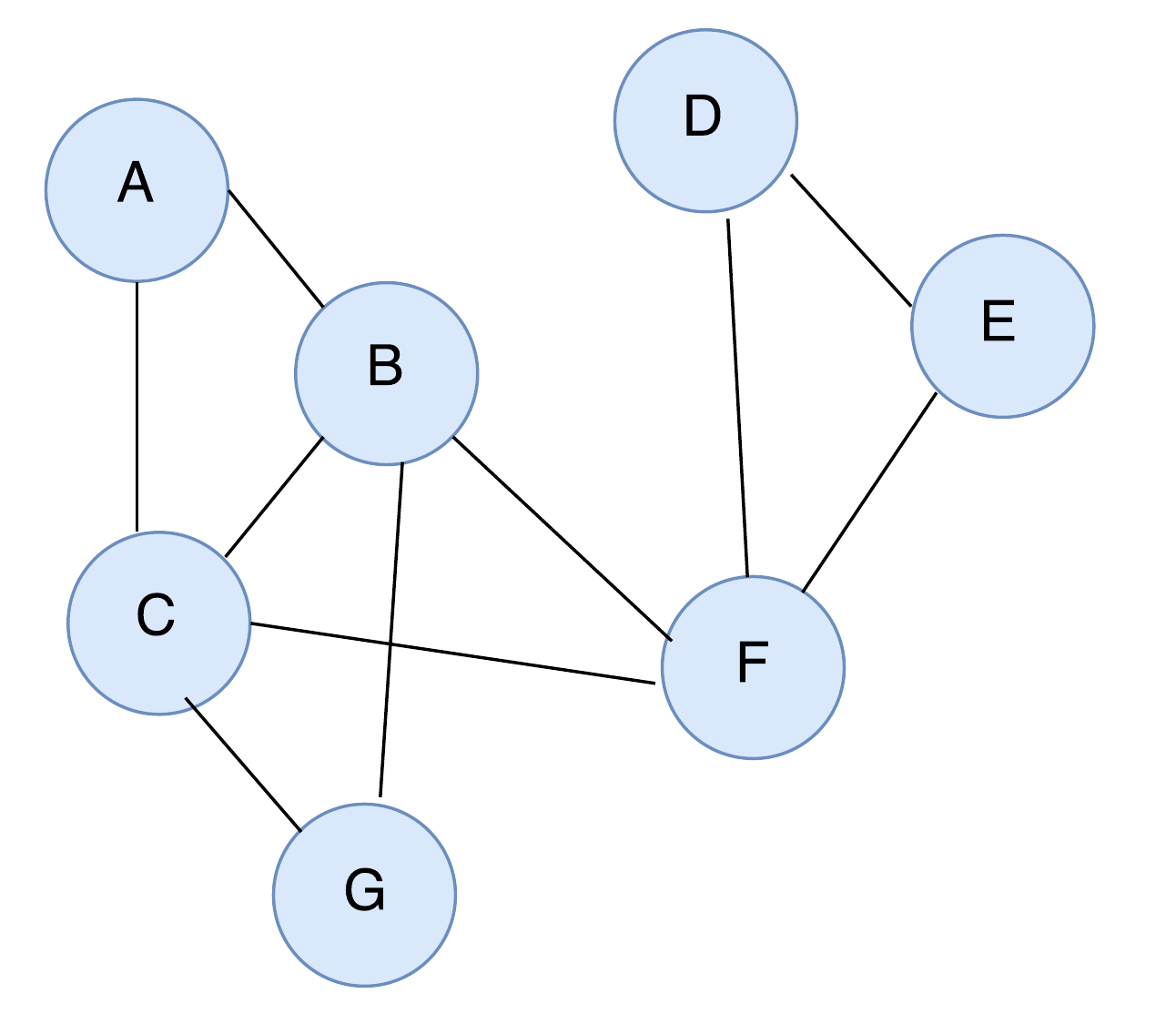}
 \end{subfigure}%
 \begin{subfigure}{0.05\textwidth}
\centering
\includegraphics[width=\linewidth]{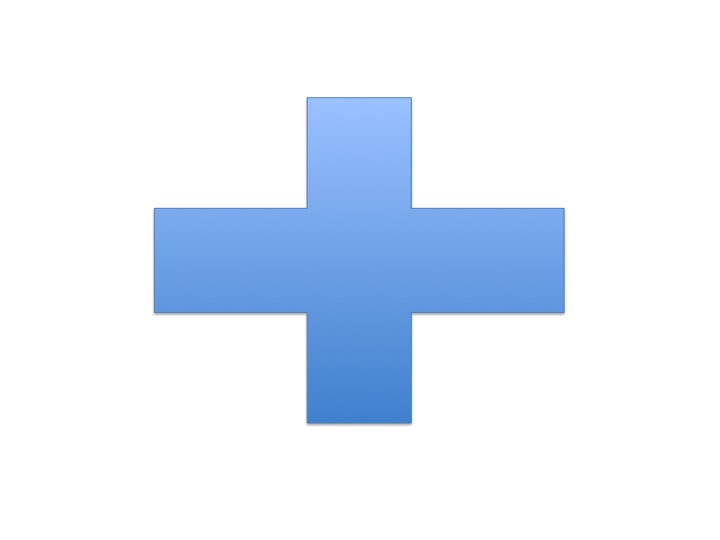}
 \end{subfigure}%
 \begin{subfigure}{0.16\textwidth}
\centering
\includegraphics[width=\linewidth]{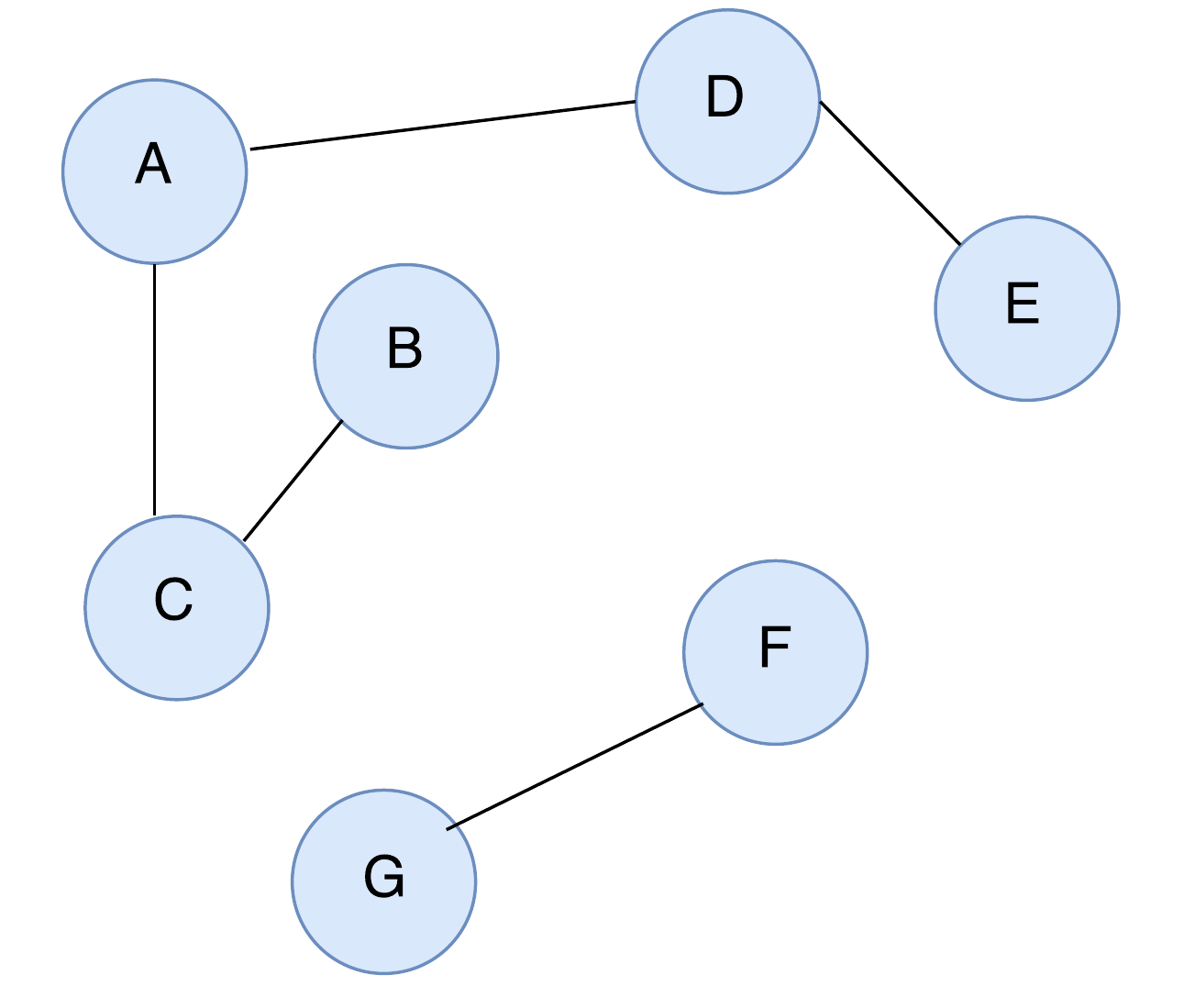}
 \end{subfigure}%
  \begin{subfigure}{0.05\textwidth}
\centering
\includegraphics[width=\linewidth]{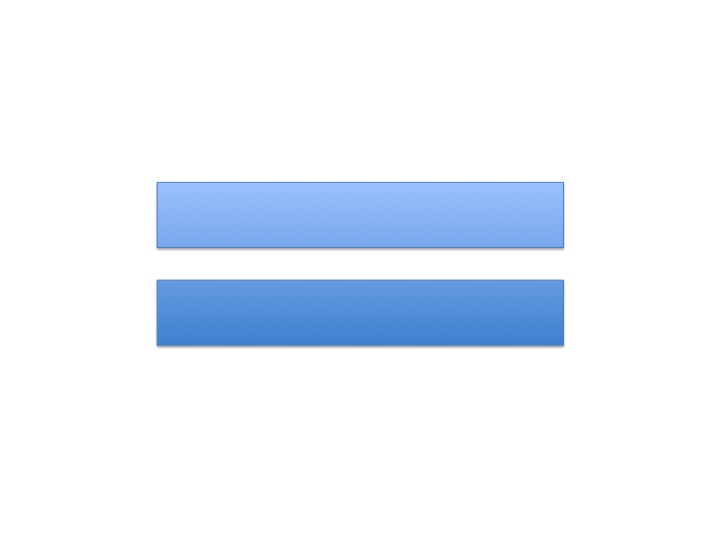}
 \end{subfigure}%
 \begin{subfigure}{0.16\textwidth}
\centering
\includegraphics[width=\linewidth]{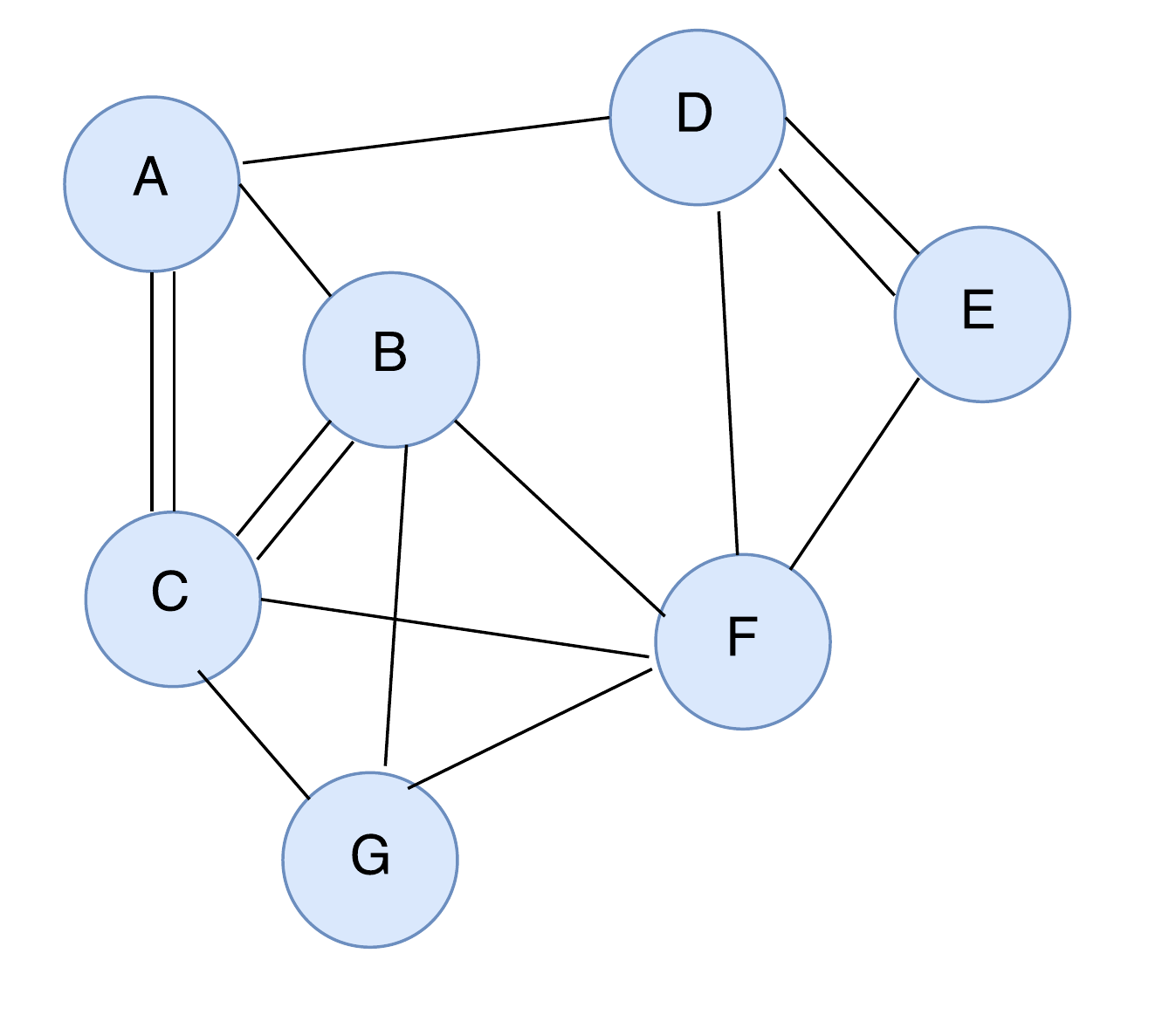}
 \end{subfigure}
 \begin{center}
     (a) \hspace{70pt} (b) \hspace{70pt} (c) \hspace{70pt} (d)
 \end{center}
 \caption{(a) A realization of the graph $G_t$ with $n=7$ vertices, before multi-edge collapsing; (b) the collapsed graph $G_t$; (c) the dyadic graph $G_e$, and (d) the superimposed graph $G_s$.}
 \label{superimposed}
\end{figure}

The graph generation process is depicted by an example in Figure~\ref{superimposed}. Observe that the superimposed graph is allowed to contain multi-edges (or, more precisely, exactly two edges) between two vertices if and only if those vertices are involved in both at least one triangle in $G_t$ and an edge in $G_e$. A practical justification for this choice of a multi-edge model comes from the fact that pair-wise and triple-wise affinities often provide complementary information\footnote{For example,~\cite{laniado2016gender} studied gender patterns in dyadic and triadic ties in an online social network and found different degrees of gender homophily in different types of ties. Hence instead of duplicating evidence from the same source, we retain two parallel edges in the graph only if they reinforce the information provided by each other.}. 
Clearly, the resulting graph $G_s$ has dependences among its edges and strong local clustering properties for properly chosen matrices $\mathbb{P}^t$ due to the increased presence of  triangles.

Furthermore, we would like to point out that this inhomogeneous superimposed random graph model differs in a number of important ways from non-uniform hypergraph random graph models on which the non-uniform hypergraph SBM, analyzed by~\cite{ghoshdastidar2017consistency,chien2018community} and others, is based. First, our model captures networks in which we cannot differentiate between an ``ordinary'' edge and a hyperedge, as hyperedges simply appear as higher-order structures in the graph. In contrast, the non-uniform hypergraph SBM is a model for networks in which different types of hyperedges are distinguishable during the observation process and labelled. Hence, a major technical difficulty of analyzing methods under the SupSBM is to deal with edge dependencies which are not present in the non-uniform hypergraph SBM.  Second, we collapse all multi-edges generated in the hyperedge generation process into single edges which are more realistic as observable network interaction models. We do, however, allow for double edges if there is complementary evidence of both dyadic and triadic ties.

In the simplest incarnation of the model, one may choose $(P^e)_{ij}=p^e$ for all $i,j$ and $(\mathbb{P}^t)_{ijk}=p^t$ for all $i,j,k$. In this case, the graph $G_e$ is a classical Erd\"os-R\'enyi dyadic random graph, while $G_t$ before multi-edges collapsing may be thought of as a generalization of Erd\"os-R\'enyi graphs to the triadic setting.

We describe next the superimposed stochastic block model based on $G_s$ graphs. 

\subsection{Superimposed stochastic block models}

Our superimposed stochastic block model (SupSBM) is based on the inhomogeneous superimposed random graph framework defined in the previous section. We consider two types of SupSBMs. In the first case, ``community signals'' are present both in the higher-order structures and the dyadic edges, while in the second case, the ``community signals'' are present only in the higher-order structures but not in the dyadic edges. Drawing a parallel with the classical SBM, where intra- and inter-community edges are generated via Erd\"os-R\'enyi graphs, both the intra- and inter-community edges in SupSBM are generated by superimposed random graph models ($G_s$) as defined in the previous section.

 We formally define a graph with $n$ vertices and $k$ communities generated from a SupSBM as follows. Each vertex of the graph is assigned a community label vector of length $k$, which takes the value of $1$ at the position corresponding to its community and $0$ at all other positions. To organize the labels, we keep track of an $n \times k$ community assignment matrix $C$ whose $i$th row $C_i$ is the community label vector for the $i$th vertex.  Given the community assignments for all the vertices in the graph, the triangle hyperedge indicators $T_{ijk}$ involving three distinct vertices $i$, $j$, $k$ are (conditionally) independent, and they follow a Bernoulli distribution with a parameter that depends only on the community assignments, i.e.,
\[
P(T_{ijk}=1| C_{ip} =1, C_{jq}=1, C_{kl}=1) = \pi^{t}_{pql}, \quad p,q,l \in \{1,\ldots ,k\},
\]
where $\pi^{t}$ is a $3$-way $k \times k \times k$ tensor of parameters. The triangle hyperedges naturally reduce to a triangle, and as before, multi-edges are collapsed to form the graph $G_{t}$.

An edge between two vertices $i$ and $j$ is generated independently of other edges and hyperedges following a Bernoulli distribution with a parameter that also depends on the community assignments, so that the edge indicator variable 
$E_{ij}$ satisfies
\[
P(E_{ij}=1| C_{ip} =1, C_{jq}=1) = \pi^{e}_{pq}, \quad p,q \in  \{1,\ldots ,k\},
\]
where $\pi^{e}$ is a $k \times k$ matrix of model parameters. For the case that the community structure is present only in the higher-order structures and not at the level of dyadic edges, this parameter equals $p^e$ irrespective of the communities that the vertices $i$ and $j$ belong to. 
The desired graph is obtained by superimposing $G_t$ and $G_e$ following the process described in the previous section. 

The above described model contains a large number of unknown parameters, and to enable a more tractable analysis, we proceed as follows. We define the stochastic block model on  $3-$hyperedges in the following manner:
\[
P(T_{ijk}=1| C_{i}, C_{j}, C_{k}) = \begin{cases}
\frac{a_t}{n}, &  \quad \text{ if } C_{i} = C_{j} = C_{k} \\
\frac{b_t}{n}, &  \quad \text{ otherwise}
\end{cases},
\]
so that the probability of a triangle hyperedge equals $\frac{a_t}{n}$ if the three vertices involved are in the same community, and $\frac{b_t}{n}$ if at least one of the vertices is in a different community than the other two. 
The dyadic edges are generated according to the following rule: the probability of an edge is $\frac{a_e}{n}$ if both the end points belong to the same community and $\frac{b_e}{n}$ if they belong to different communities.

Another simplification consists in assuming that all communities are of the same size, leading to balanced $n$-vertex $k$-block SupSBMs, $G_s(n,k,C,a_e,b_e,a_t,b_t)$, 
in which all the $k$ communities have $\frac{n}{k}$ vertices and the matrix $C$ is an $n \times k$ community assignment matrix.

\section{Analysis of higher-order spectral clustering} \label{sec:analysis}

Spectral clustering methods for hypergraphs, also known as higher-order spectral clustering methods, have been studied in a number of recent papers~\citep{zhou2006learning,benson2016higher,tsourakakis2017scalable,li2017inhomogoenous}. In particular,~\cite{benson2016higher} introduced a method that creates a ``motif adjacency matrix" for each motif structure of interest. In a motif adjacency matrix, the $(i,j)$th element represents the number of motifs that include the vertices $i$ and $j$. Spectral clustering is applied to the motif adjacency matrix in a standard form in order to find communities of motifs. While there are many variants of spectral clustering that may be applied to the motif adjacency matrices, throughout our analysis, we investigate only one algorithm, which computes the $k$ eigenvectors corresponding to the $k$ largest in absolute value eigenvalues of the motif adjacency matrix. The algorithm subsequently performs a $(1+\epsilon)$-approximate, $\epsilon>0$, $k$-means clustering~\citep{kumar2004simple,lei2015consistency} on the rows of the resultant $n \times k $ matrix of eigenvectors. Furthermore, we only consider two motif adjacency matrices, involving edges and triangles. 
The primary goal of our analysis is to describe how to detect the community structures of the SupSBM from observed triangle patterns using spectral clustering. We will consider both versions of SupSBM, namely, one with community structure present only at the triangle level, and another with community structure present both at the triangle and edge levels. In what follows, we first prove a number of concentration results for certain motif adjacency matrices under the more general inhomogeneous superimposed random graph model. Subsequently, we specialize our analysis to the SupSBMs.  

\subsection{Higher-order spectral clustering and superimposed random graphs}

\begin{figure}[h]
\centering
\vspace{-15pt}
\includegraphics[width=0.95\linewidth]{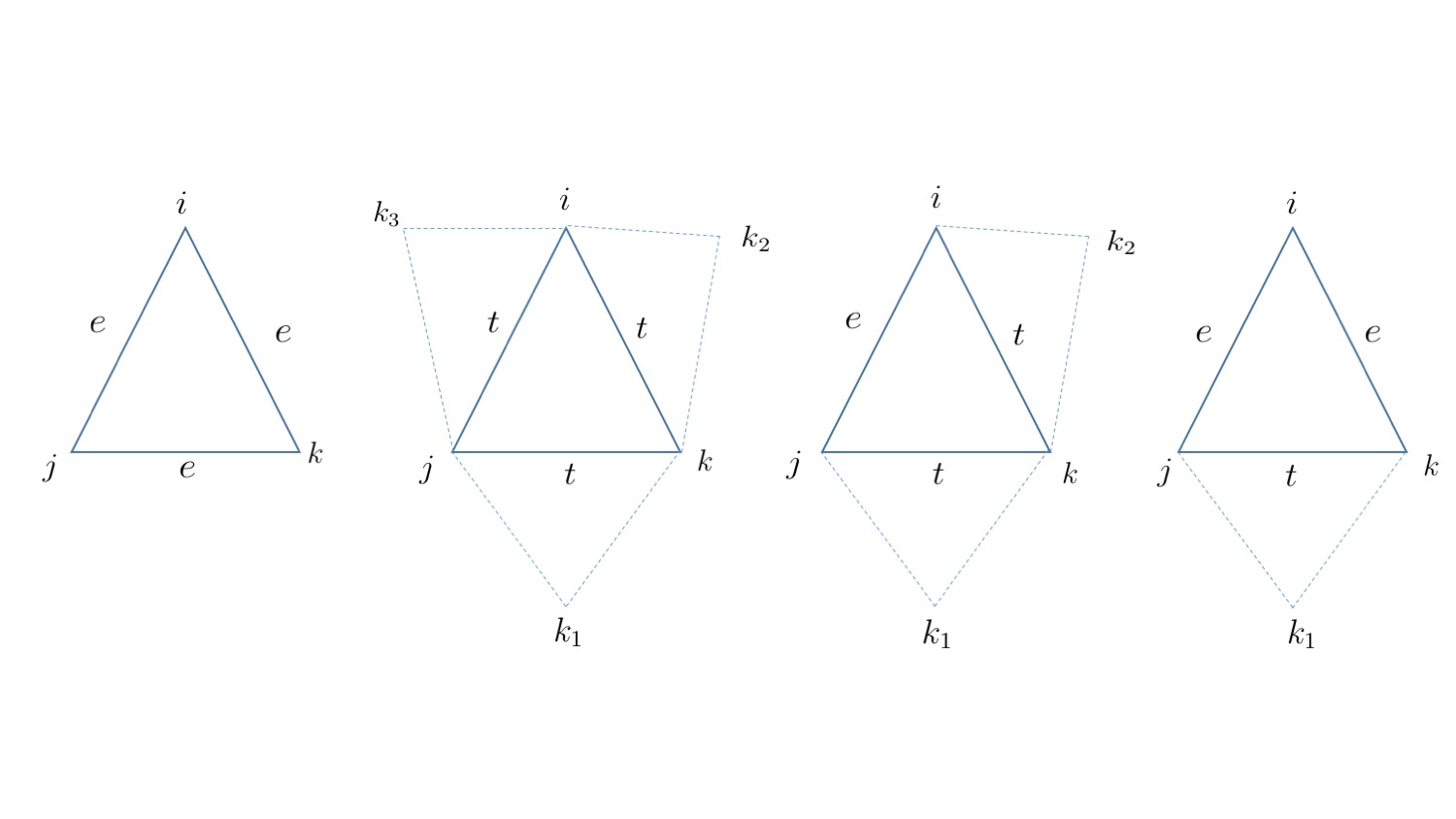}
\vspace{-40pt}
 \begin{center}
     (a) \hspace{80pt} (b) \hspace{80pt} (c) \hspace{80pt} (d)
 \end{center}
 \caption{Imposed triangles generated through the superimposition of edges and triangles: (a) $E^3$, (b) $T^3$, (c) $T^2E$, and (d) $TE^2$.} 
 \label{incidental}
\end{figure}

Let $G \sim G_s(n,P^e,P^t)$ be a graph generated from the inhomogeneous superimposed edge-triangle random graph model. We introduce two matrices, $A_E$ and $A_T$, respectively; $(A_E)_{ij}$ represent the number of observed edges between the vertices $i$ and $j$, while $(A_T)_{ij}$ represents the number of observed triangles including both $i$ and $j$ as vertices. Note these matrices are not the motif adjacency matrices of $G_e$ and $G_t$, since there are edges in $G_t$ that contribute to $A_E$ and triangles from $G_e$ that contribute to $A_T$.  There may be many ``incidentally generated" or imposed triangles~\citep{chandrasekhar2014tractable} that arise due to superimposition which also contribute to $A_T$. The different scenarios are depicted in Figure~\ref{incidental}. For our analysis, we also introduce the following two matrices: 
\begin{itemize}
\tightlist
    \item $A_{E^2}$: the adjacency matrix of edges in $G_e$; here, $(A_{E^2})_{ij} = E_{ij}.$
    \item $A_{T^2}$: the adjacency matrix of triangle motifs in $G_t$; here, $(A_{T^2})_{ij} = \sum_{k} T_{ijk}.$
\end{itemize}

As noted in~\cite{chandrasekhar2014tractable}, there are four classes of matrices needed to describe the model, namely
  \begin{itemize}
\tightlist  
    \item[(a)] $A_{E^3}$: the motif adjacency matrix of all triangles formed by random edges from $G_e$. The generative random variable for these triangles reads as:
    \[
E^3_{ijk} = E_{ij}E_{jk}E_{ik},
\]
and $(A_{E^3})_{ij} = \sum_{k} E_{ij}E_{jk}E_{ik}$.
    \item[(b)] $A_{T^3}$: the motif adjacency matrix of all triangles formed by three intersecting triangles from $G_t$. The generative random variable for these triangles reads as:
    \[
T^3_{ijk}=(1-T_{ijk}) 1(\sum_{k_1 \neq k}T_{ijk_1}>0)1(\sum_{k_2 \neq i}T_{jkk_2}>0)1(\sum_{k_3\neq j}T_{ikk_3}>0),
\]
and $(A_{T^3})_{ij} = \sum_{k} T^3_{ijk}$. 
    \item[(c)] $A_{T^2E}$: the motif adjacency matrix of all triangles formed by two triangles from $G_t$ and one edge from $G_e$. The generative random variable for these triangles reads as:
\begin{align*}
T^2E_{ijk} = (1-T_{ijk}) &1(\sum_{k_1 \neq k}T_{ijk_1}>0 \, \cap \, E_{ij}=0) \\
&1(\sum_{k_2\neq i}T_{jkk_2}>0 \, \cap \, E_{jk}=0)  1(\sum_{k_3\neq j}T_{ikk_3}=0 \, \cap \, E_{ik}=1),
\end{align*}
and $(A_{T^2E})_{ij} = \sum_{k} T^2E_{ijk}$. 
\item[(d)] $A_{TE^2}$: the motif adjacency matrix of all triangles formed by one triangle from $G_t$ and two edges from $G_e$. The generative random variable for these triangles reads as:
\begin{align*}
TE^2_{ijk} = (1-T_{ijk}) &1(\sum_{k_1 \neq k}T_{ijk_1}>0 \,\cap \, E_{ij}=0)\\
&1(\sum_{k_2\neq i}T_{jkk_2}=0 \, \cap \, E_{jk}=1) 1(\sum_{k_3\neq j}T_{ikk_3}=0 \,\cap \, E_{ik}=1),
\end{align*}
and $(A_{TE^2})_{ij} = \sum_{k} TE^2_{ijk}$. 
    \end{itemize}

Note that except for case (a), an imposed triangle involving the vertices $(i,j,k)$ arises only if there is no model-created triangle involving $(i,j,k)$ already present. Hence, the definitions of each of the random variables $T^3$, $T^2E$ and $TE^2$ include $(1-T_{ijk})$ as a factor that indicates this dependence. 
For case (a), since we allow a multiedge between two vertices that are both involved in a triangle hyperedge and an edge, it is possible to have an imposed triangle in addition to a model-generated triangle on the same triple of vertices.

With these definitions, we have that the triangle adjacency matrix reads as
$$A_{T}=A_{T^2}+A_{TE}+A_{T^2E}+A_{T^3}+ A_{TE^2},$$ 
capturing both model-based and imposed triangles. 
Obviously, we only observe the matrices $A_E$ and $A_T$ and not their specific constituents, as in real networks we do not have labels describing how an interaction is formed. Hence, even though the community structure is most explicitly described by $A_{T^2}$, we need to analyze how this matrix reflects on $A_T$ and what the properties of the latter matrix are based on $A_{T^2}$. Then, the expectation of $A_{T}$ equals
\[
E[A_{T}] = E[A_{T^2}]+E[A_{TE}]+E[A_{T^2E}]+E[A_{T^3}]+ E[A_{TE^2}],
\]
where all operators are used component-wise.

\subsubsection{Notation and asymptotic properties of superimposed graphs}
We start with some notation. Let
\[p^e_{\max}=\max_{i,j} p^e_{ij} \quad \text{ and } \quad  p^t_{\max}=\max_{i,j,k} p^t_{ijk},
\]
denote the maximum probability of edge inclusion in $G_e$ and triangle hyperedge inclusion in $G_t,$ respectively. As noted by~\cite{chandrasekhar2014tractable}, in the superimposed random graph framework, the generative probabilities summarized in the two matrices $P^t$ and $P^e$ must satisfy certain conditions in order to ensure that the imposed triangles do not significantly outnumber the generative triangles. Accordingly, we impose the following asymptotic growth conditions on $p^t_{\max}$ and $p^e_{\max}$: 
\begin{equation}
    c_1\frac{\log n}{n} \leq p^e_{\max}<c_2\frac{n^{2/5-\epsilon}}{n},
    \label{pemax}
\end{equation}
\begin{equation}
    c_1\frac{(\log n)^8}{n^2}<p^t_{\max}<c_2\frac{n^{2/5-\epsilon}}{n^2},
    \label{ptmax}
\end{equation}
 and $p^t_{\max} > c_3 p^e_{\max} \frac{\log n}{n}$ for some $\epsilon>0$ and constants $c_1$, $c_2$, and $c_3$ independent of $n$. A typical example are the following two growth rates:  $p^e_{\max}=O(\frac{\log n}{n})$ and $p^t_{\max}=O(\frac{n^{1/4}}{n^2})$. 
Note that the asymptotic growth bounds are required only for the analysis of superimposed random graphs under the SupSBM. We do not require these relations to hold for results regarding regular SBMs or 3-uniform hypergraph SBMs. 
Hence, we will not make any assumptions on the asymptotic growth bounds for $p^t_{\max}$ and $p^e_{\max}$ until Theorem~\ref{AT}.  

The following five results, summarized in Theorems \ref{ATT} to \ref{ATEE}, provide non-asymptotic error bounds that hold in more general settings, as described in the statements of the respective theorems. Note that we make repeated use of the symbols $c$ or $r$ to represent different generic constants as needed in the proofs in order to avoid notational clutter.

It is well-known that the Frobenius norm $\|A_{E^2}-P_{E^2}\|_2$ is bounded by $c_1\sqrt{\Delta}$ with probability at least $1-n^{-r}$ \citep{lei2015consistency,gao2015achieving,chin2015stochastic}, where $\Delta = \max\{np^e_{\max},c\log n\}$. The following theorems establish similar upper bounds for other component matrices involved in our analysis as well as a bound on $\|A_{T}-P_{T}\|_2$. The proofs of all theorems are delegated to the Appendix.

\subsubsection{Concentration bounds for $\mathbf{A_{T^2}}$}

\begin{thm}
Let $G_t(n,\mathbb{P}^t)$ be a $3$-uniform hypergraph in which each possible 3-hyperedge is generated according to a Bernoulli random variable $T_{ijk}$ with parameter $p^t_{ijk},$ independent of all other 3-hyperedges. Let $A_{T^2}$, as before, stand for the triangle-motif adjacency matrix. Furthermore, let $\Delta_{t} = \max \{n^2 p^t_{\max},c \log n\}$. Then, for some constant $r>0$, there exists a constant $c_1(c,r)>0$ such that with probability at least $1-n^{-r}$, one has
\[
\|A_{T^2}-E[A_{T^2}]\|_2 \leq c_1\sqrt{\Delta_{t}}.
\]
\label{ATT}
\end{thm}
Note that in the above bound, $\Delta_{t}$ may be interpreted as being an approximation of the maximum expected ``triangle degree'' of vertices in $G_t$. Drawing a parallel with adjacency matrices of graphs, one may define the ``degree" of a row of an arbitrary matrix as the sum of the elements in that row. Then, $\Delta_t$ is an upper bound on the degree of a row in the matrix $A_{T^2},$ much like $\Delta$ is an upper bound for the degrees of the rows in $A_{E^2}$. The above result for triangle-motif adjacency matrices is hence an analogue of a similar result for standard adjacency matrices described in ~\citet{lei2015consistency,gao2015achieving,chin2015stochastic}. The arguments used to prove the result in the cited papers are based on an $\epsilon-$net analysis of random regular graphs laid out in~\cite{friedman1989second,feige2005spectral}. We extend these arguments to the case of triangle hyperedges; due to the independence of the random variables corresponding to the hyperedges involved in all sums of interest, we do not require new concentration inequalities to establish the claim. This is not the case for the results to follow.

\subsubsection{Concentration bounds for $\mathbf{A_{E^3}}$}
Next, we derive an upper bound for the spectral norm of the matrix $A_{E^3}-E[A_{E^3}]$.  Note that the elements of the matrix $A_{E^3}$ are dependent and consequently, the sums of the random variables used in the $\epsilon-$net approach include dependent variables. Hence, the $\epsilon-$ net approach cannot be applied directly, and several substantial modifications in the proofs are needed. However, each element of $A_{E^3}$ is a low-degree polynomial of the generic independent random variables $E_{ij}$. Therefore, we show that in all the sums of dependent random variables of our interest, the dependencies between the random variables are limited, and the number of co-dependent random variables are significantly smaller than that of all the variables in the sum. In what follows, we build upon recent advances in concentration inequalities for functions of independent random variables~\citep{warnke2016method} and sums of dependent random variables \citep{warnke2017upper} to derive concentration bounds for $A_{E^3}$.

Let $\tau_{\max}=\max\{n(p^e_{\max})^2, c\log n\}$, $\Delta_{E^3}=\max\{n^2(p^e_{\max})^3, c(\log n)^2\}$ and $D_{E^3} = np^e_{\max}\tau_{\max}^2=\max \{n^3 (p^e_{\max})^5, c\,np^e_{\max}(\log n)^2\}$ and assume $np^e_{\max} \geq c \log n$. We have the following result.
\begin{thm}
Let $G_e(n,P^e)$ be an inhomogeneous random graph in which each edge is independently generated by a Bernoulli random variable  $E_{ij}$  with parameter $p^e_{ij}, \, i,j=1,\ldots,n$. Let $D_{E^3} = \max \{n^3 (p^e_{\max})^5, cnp^e_{\max}(\log n)^2\}$ for some constant $c$. Then, for some constant $r>0$, there exists a constant $c_2(c,r)>0$ such that with probability at least $1-n^{-r}$,
\[
\|A_{E^3} - E[A_{E^3}]\|_2 \leq c_2 \sqrt{D_{E^3}}.
\]
\label{ATE}
\end{thm}

\begin{figure}[h]
\centering
\begin{subfigure}{0.3\linewidth}
\includegraphics[width=\linewidth]{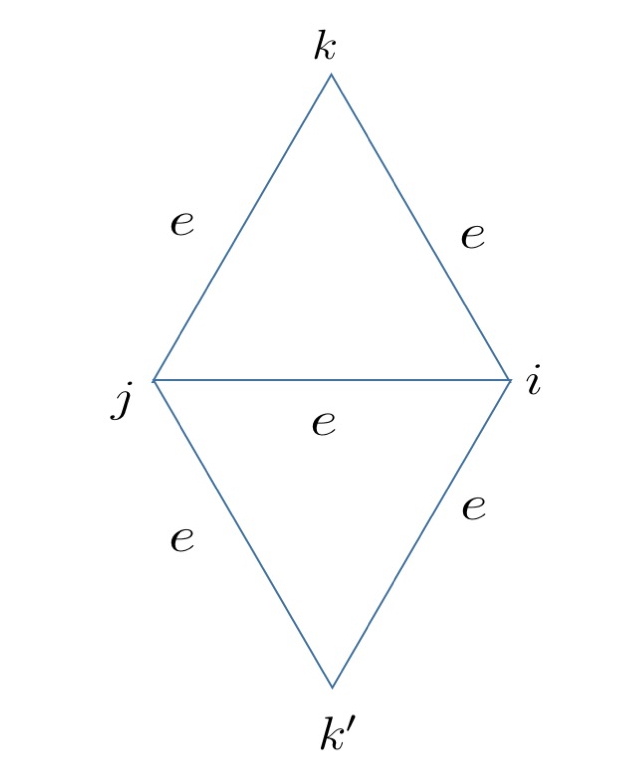}
 \end{subfigure}%
 \begin{subfigure}{0.3\linewidth}
\includegraphics[width=\linewidth]{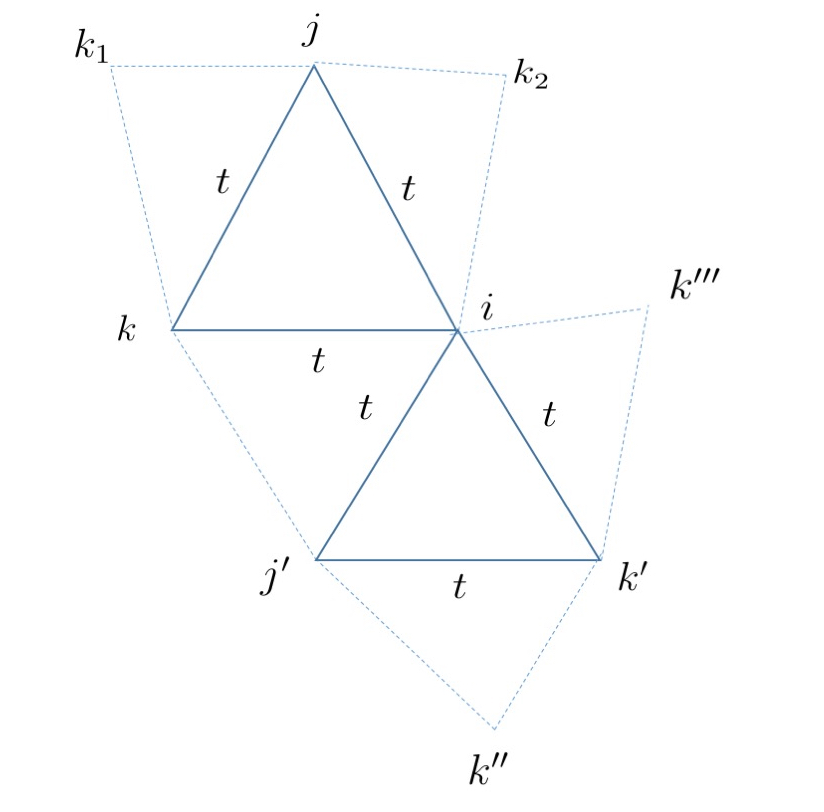}
 \end{subfigure}%
 \begin{subfigure}{0.3\linewidth}
\includegraphics[width=\linewidth]{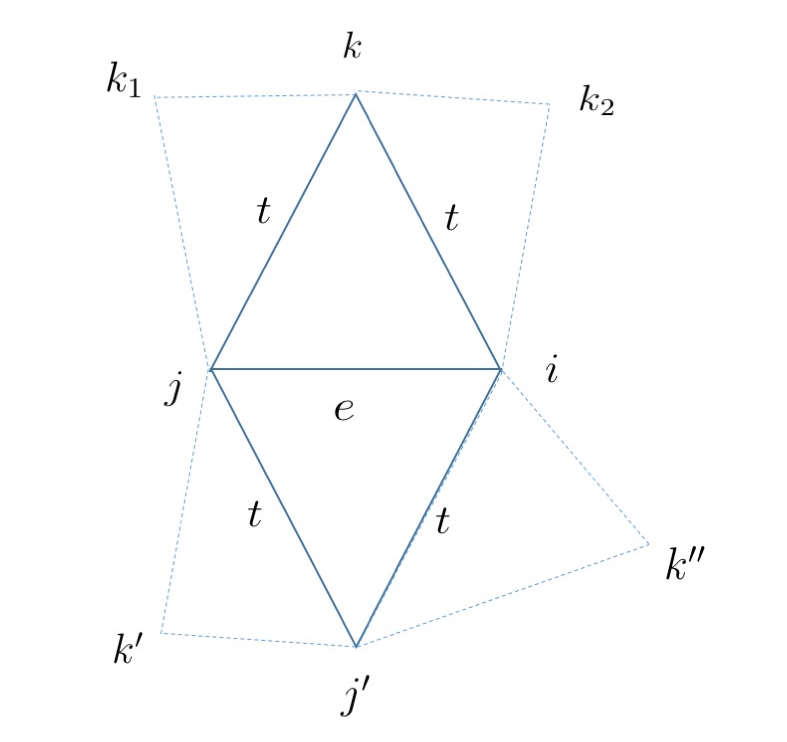}
 \end{subfigure}
  \begin{center}
     (a) \hspace{100pt} (b) \hspace{100pt} (c)
 \end{center}
\begin{subfigure}{0.3\linewidth}
\includegraphics[width=\linewidth]{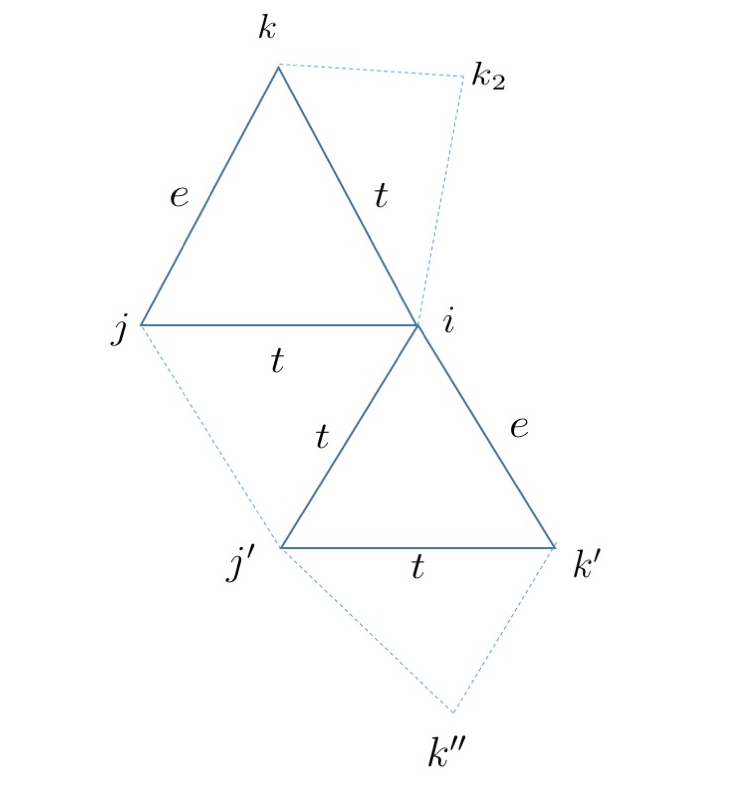}
 \end{subfigure}%
 \begin{subfigure}{0.3\linewidth}
\includegraphics[width=\linewidth]{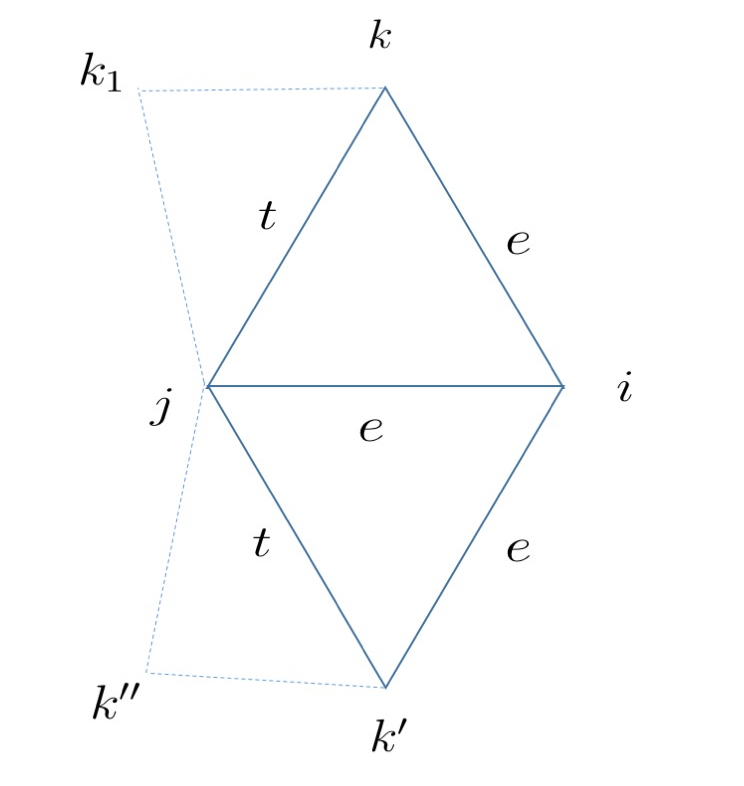}
 \end{subfigure}%
 \begin{subfigure}{0.3\linewidth}
\includegraphics[width=\linewidth]{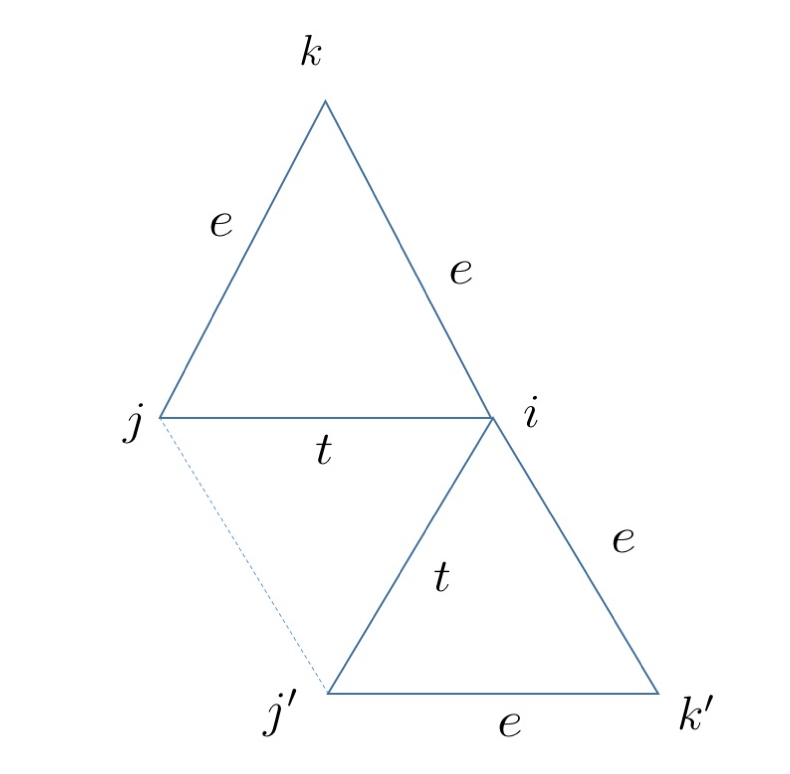}
 \end{subfigure}
  \begin{center}
     (d) \hspace{100pt} (e) \hspace{100pt} (f)
 \end{center}
 \caption{Dependence among the random variables of incidental triangles that include vertex $i$, (a) $E^3$ (b) $T^3$ (c) $T^2E$ of type 1, (d) $T^2E$ of type 2, (e) $TE^2$ of type 1, and (f) $TE^2$ of type 2.}
 \label{dependence}
\end{figure}

The proof of the result is based on Theorem 1.3 of~\cite{warnke2016method} and Theorem 9 of~\cite{warnke2017upper}. We still resort to the use of $\epsilon-$nets but also take into account the particular dependencies between the random variables. The key observations are that two triangles generated by edges from $G_e$ are dependent if they share an edge (see Figure \ref{dependence}(a)), and, with high probability, each triangle shares an edge with at most $2\tau_{\max}$ other triangles. The random variables whose sums we are interested in bounding represent such triangles. Note that our result is a stronger bound than the one actually needed for obtaining an upper bound for  $\|A_{T} - P_{T}\|_2 $ under the asymptotic growth conditions of interest. Indeed, Proposition \ref{prop1} stated in the Appendix automatically gives an upper bound of the form $O(\Delta_{E^3})$ for $\|A_{E^3} - E[A_{E^3}]\|_2 $. While this loose bound would have been sufficient, we resorted to a more careful analysis using the $\epsilon-$net approach and the results of Theorem 1.3 of~\cite{warnke2016method} to arrive at a significantly improved bound $O(\sqrt{D_{E^3}})$. We also note that based on the results derived for $A_{E^2}$ and $A_{T^2}$, the bound for the case when the elements of $A_{E^3}$ are mutually independent should read as $O(\sqrt{\Delta_{E^3}})$. The current bound is worse than this bound by a factor of $\sqrt{\Delta}$, and it is not immediately clear how the latter bound can be improved further.

\subsubsection{Concentration bounds for other relevant matrices}

For the next three results, we use the following property of the spectral norm of a square symmetric matrix. For any $n \times n$ square symmetric matrix $X$, define the spectral norm of $X$ as $\|X\|_2 = \sigma_{\max}(X)$, the largest singular value of $X$, the 1-norm as $\|X\|_1 = \max_{j}\sum_{i}|X_{ij}|$, and the $\infty-$norm as $\|X\|_{\infty}=\max_{i}\sum_{j}|X_{ij}|$.  Now assume $X$ is an $n \times n$ symmetric matrix whose elements are non-negative random variables. Let the entries of its expectation, $E[X]$, also be  non-negative. Then,
\begin{align}
\|X-E[X]\|_2 & \leq \sqrt{\|X-E[X]\|_1 \|X-E[X]\|_{\infty}} = \|X-E[X]\|_1 
\nonumber\\
& = \max_{i}\sum_{j}|X_{ij}-E[X]_{ij}| \leq \max_{i}\sum_{j}X_{ij} + \max_{i}\sum_{j}E[X]_{ij},
\label{twonorm}
\end{align}
where the first inequality is Corollary 2.3.2 in~\cite{golub2012matrix}, and the second equality follows since $X-E[X]$ is a symmetric matrix by assumption.
Note the first term in the final sum is the \emph{degree} of row $i$ of the matrix $X$. Hence, a high-probability bound on the maximum degree will allow us to upper bound this quantity. The second term equals the maximum expected degree of $X$, which is a deterministic quantity. For the next three theorems in this section we assume $n^2p^t_{\max} > c_0(\log n)^2$ and $np^e_{\max} > c_0' \log n$.

\begin{thm}
Let $G\sim G_s(n,P^e,\mathbb{P}^t)$ be a graph generated by the superimposed random graph model. Let 
$$\Delta_{T^3} = \max \{n^5(p^{t}_{\max})^3,c(\log n)^4\},$$ 
where $c$ is some constant. Then, there exists a constant $c' > 0$, such that with probability at least $1-o(1)$, one has
\[
\|A_{T^3}- E[A_{T^3}]\|_2 \leq c'\, \Delta_{T^3}.
\]
\label{ATTT}
\end{thm}
 
\begin{thm}
Let $G\sim G_s(n,P^e,\mathbb{P}^t)$ be a graph generated by the superimposed random graph model. Let 
$$\Delta_{T^2E} = \max \{n^4(p^{t}_{\max})^2p^e_{\max},c(\log n)^4\},$$ 
where $c$ is some constant. Then, there exists a constant $c' > 0$, such that with probability at least $1-o(1)$, one has
\[
\|A_{T^2E}- E[A_{T^2E}]\|_2 \leq c'\, \Delta_{T^2E}.
\]
\label{ATTE}
\end{thm}

\begin{thm}
Let $G\sim G_s(n,P^e,\mathbb{P}^t)$ be a graph generated by the superimposed random graph model. Let 
$$\Delta_{TE^2} = \max \{n^3p^{t}_{\max}(p^e_{\max})^2,c (\log n)^3\},$$ 
where $c$ is some constant. Then, there exists a constant $c' > 0$, such that with probability at least $1-o(1)$, one has
\[
\|A_{TE^2}- P_{TE^2}\|_2 \leq c'\, \Delta_{TE^2}.
\]
\label{ATEE}
\end{thm}

The proofs of all three above results follow a similar outline. In each case, the degree of a row $i$ is a sum of \textit{dependent} triangle-indicator random variables for triangles that include vertex $i$. 
However, in each case we carefully enumerate the events that lead to two such incidental triangle indicator random variables to be dependent and show that the number of such cases is limited with high probability. This allows us to apply Theorem 9 of~\cite{warnke2017upper} in an iterative manner and obtain concentration results on the respective sums. While we relegate technically involved rigorous proofs to the Appendix, we graphically illustrate all the events that lead to the dependencies that we need to consider in Figure \ref{dependence}. For the family of random variables $\{(T^3)_{ijk}, \, j,k \in \{1,\ldots,n\}\}$, two variables are dependent if and only if one of the triangles from $G_t$ covering $(T^3)_{ijk}$ has an edge $ij'$ or $ik'$ and consequently also covers $(T^3)_{ij'k'}$ (see Figure~\ref{dependence}(b)). For the family of random variables $\{(T^2E)_{ijk}, \, j,k \in \{1,\ldots,n\}\}$, two variables are dependent if either one of the edges $ij$ or $ik$ is covered by a triangle from $G_t$ and the same triangle edge-intersects $(T^2E)_{ij'k'}$ (see Figure~\ref{dependence}(c)), where one of $ij$ or $ik$ is an edge in $G_e$ and is also an edge in $(T^2E)_{ij'k'}$ (see Figure~\ref{dependence}(d)). Finally, for the family of random variables $\{(TE^2)_{ijk}, \, j,k \in \{1,\ldots,n\}\}$, two variables are dependent if either one of the edges $ij$ or $ik$ is covered by a triangle from $G_t$ and the same triangle edge-intersects $(TE^2)_{ij'k'}$ (see Figure~\ref{dependence}(e)), or, one of the two edges $ij$ and $ik$ is generated by $G_e$ and is also an edge covered by $(TE^2)_{ij'k'}$ (see Figure~\ref{dependence}(f)). 

We also note that the bounds in the previous three results can be improved by applying the $\epsilon-$net approach for dependent random variables as used in Theorem~\ref{ATE}. However, the above stated upper bounds suffice to obtain the desired concentration bound for $A_T$, as summarized in the following subsection. 

\subsubsection{Concentration bound for $A_{T}$}
In the next theorem we combine the previous results to arrive at a concentration bound for the matrix $A_T$ under the assumptions made on $p^e_{\max}$ and $p^t_{\max}$ in (\ref{pemax}) and (\ref{ptmax}).
\begin{thm}
Let $A_{T}$ denote the triangle-motif adjacency matrix of a random graph $G$ generated by the inhomogeneous superimposed random graph model $G_s(n,P^e,\mathbb{P}^t)$. If $\Delta_t >  c (\log n)^8$ for some constant $c$, and the assumptions (\ref{pemax}) and \ref{ptmax}) on $p^e_{\max}$ and $p^t_{\max}$ hold, then with probability at least $1-o(1)$, one has
\[
\|A_{T}-E[A_{T}]\|_2 \leq c' \sqrt{\Delta_{t}},
\]
\label{AT}
where $c'$ is a constant independent of $n$.
\end{thm}
We note the similarity of the upper bound of this concentration inequality with that obtained for $A_{T^2}$ in Theorem \ref{ATT}. The above result then tells us that the effect of the incidental triangles on the concentration of $A_T$ is limited and the rate in the upper bound is predominantly determined by the rate for $A_{T^2}$. This suggests that while the superimposition process induces dependencies between the edges in $G_s$ through the presence of triangles from $G_t$, the model, under suitable sparsity conditions, is still mathematically tractable. The influence of the incidental triangles can be analyzed and controlled.  

Next, we turn our attention to analyzing random graphs generated by SupSBMs, and focus in particular on quantifying the misclustering error rate under a higher-order spectral clustering algorithm.  

\subsection{Higher-order spectral clustering under the SupSBM}

Let $G \sim G_s(C,n,k,a_e,b_e,a_t,b_t)$ be a graph generated by the balanced $n$-vertex, $k$-block SupSBM with a community assignment matrix $C$ as defined before. Let $\hat{C}$ denote the $n \times k$ matrix of eigenvectors corresponding to the $k$ largest absolute-value eigenvalues of the triangle motif adjacency matrix $A_{T}$. To obtain the community assignments for the vertices, a $(1+\epsilon)$-approximate $k$-means clustering, with $\epsilon>0$, is performed on the rows of $\hat{C}$~\citep{kumar2004simple,lei2015consistency}.

We define the misclustering error rate $R$ as follows. Let $\bar{e}$ and $\hat{e}$ denote the vectors containing the true and estimated community labels of all the vertices in $V$. Then we define
 \[
 R = \inf_{\Pi} \frac{1}{n}  \sum_{i=1}^{n} 1( \bar{e}_i \neq \Pi(\hat{e}_i)),
 \]
where the infimum is taken over all permutations $\Pi(\cdot)$ of the community labels. 

To bound the misclustering rate $R$, one needs to relate it to the difference between the estimated and the true eigenvectors. For this purpose, one can use the well-known Davis-Kahan Theorem~\citep{dk70,stewart} that characterizes the influence of perturbations on the spectrum of a matrix. For a symmetric matrix $X$, let $\lambda_{\min}(X)$ stand for its smallest in absolute value non-zero eigenvalue. Since $\hat{C}_{n \times k}$ 
 
is the matrix of eigenvectors it has orthonormal columns, and hence we have the following bound
\begin{equation}
R \leq \frac{1}{n}\frac{n}{k}8 (2+\epsilon)\|\hat{C}-C(C^TC)^{-1/2}\mathcal{O}\|_F^2 \leq 64(2+\epsilon)\frac{\|A_T-E[A_T]\|_2^2}{(\lambda_{\min}(E[A_T])^2} ,
\label{misclus}
\end{equation}
where $\mathcal{O}$ is an arbitrary orthogonal matrix~\citep{lei2015consistency} and the last inequality arises from the Davis-Kahan Theorem.

Next, we derive a lower bound on $\lambda_{\min}(E[A_T])$. We start by computing the expectations of the motif adjacency matrices $A_{E^2}$, $A_{T^2}$, and $A_{E^3}$ under the SupSBM. In all three cases, these expectations are of the form $C((g-h)I_k + h1_k1_k^T)C^T$, where as before $C$ denotes the community assignment matrix, $I_k$ is the $k$-dimensional identity matrix, $1_k$ is the $k$-dimensional vector of all $1$s, and $g$ and $h$ are functions of the parameters $n,k,a_e,b_e,a_t,b_t$.

 For matrices of the form $C((g-h)I_k + h1_k1_k^T)C^T$, with $g>h>0$, $1_k$ is an eigenvector corresponding to the eigenvalue $\frac{n}{k}(g-h)+nh,$ and the remaining non-zero eigenvalues are of the form $\frac{n}{k}(g-h),$ where the values of $g$ and $h$ differ for the different matrices~\citep{rcy11}. Since $nh>0$, the smallest non-zero eigenvalue equals $\frac{n}{k}(g-h)$.

We start by analyzing $A_{E^2}$. Clearly, 
\begin{align*}
E[A_{E^2}]= C\left(\frac{(a_e-b_e)}{n}I_k + \frac{b_e}{n}1_k1_k^T\right)C^T,
\end{align*}
so that $\lambda_{\min}(E[A_{E^2}])=\frac{a_e-b_e}{k}.$

Next, we note that the expected value of $A_{T^2}$ equals $E[A_{T^2}]_{ij} = \sum_{k \neq i,j} p^t_{ijk}$.
When $C_i = C_j$, i.e., when the vertices $i$ and $j$ are in the same community, then  
\[
E[A_{T^2}]_{ij} =\left(\frac{n}{k}-2\right) \frac{a_t}{n} + (k-1)\frac{n}{k}\frac{b_t}{n},
\]
while when $C_i \neq C_j,$  
\[
E[A_{T^2}]_{ij} = (n-2) \frac{b_t}{n}.
\]
The difference between the two above entities equals
\[
\left(\frac{n}{k}-2\right) \frac{a_t}{n} + (k-1)\frac{n}{k}\frac{b_t}{n}-(n-2) \frac{b_t}{n} = \left(\frac{n}{k}-2\right) \frac{a_t-b_t}{n}.
\]
Hence,
\[
E[A_{T^2}] = C \left(
\left(\frac{n}{k}-2\right) \frac{a_t-b_t}{n}I_k + (n-2) \frac{b_t}{n}  1_k1_k^T \right)C^T.
\]

Consequently,
\begin{equation}
    \lambda_{\min}(E[A_{T^2}]) = \frac{n}{k} \left(\frac{n}{k}-2\right) \frac{a_t-b_t}{n} = \left(\frac{n}{k}-2\right) \frac{a_t-b_t}{k}.
    \label{lambdaPTT}
\end{equation}
To determine $E[A_{E^3}]$, we first note that
\[
   E[A_{E^3}]_{ij} = \sum_{k \neq i,j} p_{ij}p_{jk}p_{ik} = p_{ij} \sum_{k \neq i,j} p_{jk}p_{ik}.
\]
When $C_i =C_j$, 
\[
E[A_{E^3}]_{ij} = \frac{a_e}{n} \,\{{ \left(\frac{n}{k}-2\right) \frac{a^2_e}{n^2} + (k-1)\frac{n}{k} \frac{b^2_e}{n^2}\}},
\]
while when $C_i \neq C_j$, 
\[
E[A_{E^3}]_{ij} = \frac{b_e}{n}\, \{ 2\left(\frac{n}{k}-1\right) \frac{a_eb_e}{n^2} + (k-2)\frac{n}{k} \frac{b_e^2}{n^2}\}.
\]
The difference between the above two probabilities equals 
\begin{align*}
    \frac{b_e^2(a_e-b_e)}{n^2} + \frac{\left(a_e^2+a_eb_e-2b_e^2\right)(a_e-b_e)}{kn^2} - 2\frac{a_e(a_e+b_e)(a_e-b_e)}{n^3}.
\end{align*}
Hence,
\begin{align*}
E[A_{E^3}] = Z (
& (\frac{b_e^2(a_e-b_e)}{n^2} + \frac{\left(a_e^2+a_eb_e-2b_e^2\right)(a_e-b_e)}{kn^2} - 2\frac{a_e(a_e+b_e)(a_e-b_e)}{n^3})I_k \\
& \quad +  \frac{b_e}{n} ( 2(\frac{n}{k}-1) \frac{a_eb_e}{n^2} + (k-2)\frac{n}{k} \frac{b_e^2}{n^2}) 1_k1_k^T)Z^T.
\end{align*}

Consequently, the smallest non-zero eigenvalue equals
\begin{align}
    \lambda_{\min}(E[A_{E^3}]) & =  \frac{(kb_e^2 + a_e^2+a_eb_e-2b_e^2)(a_e-b_e)}{k^2n} - 2\frac{a_e(a_e+b_e)(a_e-b_e)}{kn^2}.
    \label{lambdaPTE}
\end{align}

A special case of the SBM that is widely analyzed in the literature is the balanced SBM, in which $2n$ vertices are partitioned into two blocks. In our setting, balancing implies that the SupSBM model of interest has parameters $G_s(Z,2n,2,2a_e,2b_e,2a_t,2b_t)$, and it results in $\lambda_{\min}(E[A_{T^2}])=(n-2)(a_t-b_t)$ and $\lambda_{\min}(E[A_{E^3}])=\frac{a_e(a_e+b_e)(a_e-b_e)}{n} - 2\frac{_e(a_e+b_e)(a_e-b_e)}{n^2}$.

 We are now ready to state the main result of the paper.

\begin{thm}
Let $G \sim G_s(C,n,k,a_e,b_e,a_t,b_t)$ be a graph generated from the balanced $k$ block SupSBM. If $\Delta_t>c(\log n)^6,$ then with probability at least $1-o(1)$, the misclustering rate of community detection using the higher-order spectral clustering method satisfies
\[
R_T \leq \frac{64(2+\epsilon)c_6\Delta_t}{((\frac{n}{k}-2)(a_t-b_t))^2}.
\]

\label{RT}
\end{thm}

Under the assumed growth rate on $p^t_{\max}$ we have $n^2p^t_{\max}>c(\log n)^6$, and hence we can replace $\Delta_t$ by $n^2p^t_{\max}$. Under the $k$ block balanced SupSBM model, $p^t_{\max} =\frac{a_t}{n}$. Hence ignoring the constants, we can rewrite the upper bound as
\[
R_T \lesssim  \frac{k^2a_t}{n(a_t-b_t)^2}.
\]
As an example, if we assume $a_t = m_t n^{1/4}$ and $b_t = s_t n^{1/4}$ for constants $m_t$ and $s_t$, then the result implies that it is possible to detect the communities consistently as long as $(m_t-s_t)^2 = \omega (\frac{k^2}{n^{5/4}})$.

For the special case of a $2n$-vertices and $2$-block SupSBM $G_s(C,2n,2,2a_e,2b_e,2a_t,2b_t)$, we can further simplify this upper bound to 
\begin{equation}
R \lesssim  \frac{a_t}{n(a_t-b_t)^2}.
\label{Rsup}
\end{equation}
For comparison, we note that the result in~\cite{lei2015consistency} for the misclustering rate of spectral clustering with classical adjacency matrices under the SBM reads as
$R_E \lesssim \frac{a_e}{(a_e-b_e)^2}$.

\subsection{Uniform and non-uniform hypergraph SBMs}

In what follows, we analyze the performance of the higher-order spectral clustering under the uniform and non-uniform hypergraph SBMs \citep{ghoshdastidar2017consistency,chien2018community}. The balanced $n$-vertex $k$-block \emph{3-uniform hypergraph} SBM $G_t(C,n,k,a_t,b_t)$ is defined in the following way. All the $k$ communities have an equal number of vertices $s=\frac{n}{k}$, and the probability of forming a triangle hyperedge equals $\frac{a_t}{n}$ if all three vertices belong to the same community, while the probability of forming a triangle hyperedge equals $\frac{b_t}{n}$ if one of the vertices belongs to a different community than the other two.

Non-uniform hypergraphs involve two types of hyperedges, edges and triangles, that need to be described separately. Note that this model differs from the superimposed random graph framework used throughout the paper as the observations are of the form of two-way and three-way interactions between entities. Hence, we have a way to differentiate between an edge and a triangle hyperedge. The $n$-vertex $k$-block balanced non-uniform hypergraph SBM $G_H(C,n,k,a_e,b_e,a_t,b_t)$ is defined in the same way as a SupSBM, except that we do not replace the generated triangle hyperedges with three ordinary edges and we do not collapse multiedges.   

If we assume our observed graph is generated from a uniform hypergraph SBM on triangle hyperedges, then spectral clustering of the motif adjacency matrix is equivalent to spectral clustering based on $A_{T^2}$ only. 
Let $\hat{C}^{(T^2)}$ be the matrix of eigenvectors corresponding to the $k$ largest absolute eigenvalues of the matrix $A_{T^2}$. Then, using the bound for $A_{T^2}$ in Theorem~\ref{ATT} from Section 3.1.2, we arrive at the following result.
 \begin{thm}
 Let $G_t$ be a triangle hypergraph generated from the $k$-block uniform triangle hypergraph SBM with parameters $C,n,k,a_t,b_t$. Then, with probability at least $1-n^{-c}$, the misclustering rate of the community assignments obtained using the higher-order spectral clustering algorithm applied to the triangle motif adjacency matrix equals
\[
R_{T^2} \leq 64(2+\epsilon)\frac{\|A_{T^2}-E[A_{T^2}]\|_2^2}{(\lambda_{\min}(E[A_{T^2}]))^2} \leq \frac{64(2+\epsilon)c_1\Delta_t}{((\frac{n}{k}-2)(a_t-b_t))^2}.
\]
\label{RTT}
\end{thm}
 
We can simplify this upper bound under the assumption of a $2n$-vertex $2$-block triangle hypergraph SBM $G_t(C,2n,2,2a_t,2b_t)$ to $R_{T^2} \lesssim \frac{a_t}{n(a_t-b_t)^2}.$ Note that the concentration bound is smaller by a factor of $n$ when compared to the same result for spectral clustering of ordinary edge-adjacency matrix in \cite{lei2015consistency}, provided that the parameters $a_e, b_e$ and $a_t,b_t$ are comparable. Alternatively, the misclustering error rate using the triangle motif adjacency matrix for a graph generated from a triangle hypergraph SBM is better than the corresponding rate from an edge-based adjacency matrix of a graph generated from SBM as long as $a_t \gtrsim \frac{a_e}{n}$. 

The above observation has important implication for non-uniform hypergraph SBMs. To describe why this is the case, assume that we are given a non-uniform hypergraph generated from the $2n$-vertex $2$-block balanced non-uniform hypergraph SBM $G_H(C,n,k,a_e,b_e,a_t,b_t).$ The question of interest is: Given $a_e,b_e,a_t,b_t$, with $a_e \asymp b_e$ and $a_t\asymp b_t$, should one use the edge-based adjacency matrix, the triangle-based adjacency matrix, or a combination thereof? Let
\begin{equation}
a_t \asymp \frac{a_e}{\delta}, \quad a_t-b_t=m\frac{a_e-b_e}{\delta}, \quad a_e \asymp b_e,
\label{NUHsetup}
\end{equation}
so that asymptotically, the probabilities $a_e$ and $b_e$ are $\delta$-times the probabilities $a_t$ and $b_t$, while the difference between the probabilities $a_e-b_e$ is $\frac{\delta}{m}$ times that of the difference between $a_t-b_t$. Clearly, $\delta$ captures the asymptotic difference between the densities of triangle hyperedges and dyadic edges, while $m$ captures the difference in the ``communal'' qualities between these two types of hyperedges. Note that the notation for asymptotic equivalence ignores all constants.
\begin{thm}
Let $G \sim G_{H}(C,2n,2,2a_e,2b_e,2a_t,2b_t)$ be a graph generated from the non-uniform hypergraph SBM. Assume the relationships between the probabilities $a_e,b_e,a_t,b_t$ are as in Equation~(\ref{NUHsetup}). Then, spectral clustering based on a triangle adjacency matrix has a lower error rate than spectral clustering based on an edge adjacency matrix if $\frac{\delta}{m^2n} \lesssim 1,$ and a higher error rate if $\frac{\delta}{m^2n} \gtrsim 1$.
\label{tradeoff}
\end{thm}

Note that even though in practice we do not observe the quantities $m$ and $\delta$, it is possible to estimate them reliably and efficiently. To estimate $\delta$, we only need to look at the ratio of the densities of the hyperedges. The expected degree density of triangle hyperedges is $O(na_t)$, while that of edges is $O(a_e)$. This implies that 
$$\delta \asymp \frac{a_e}{a_t}=n\frac{\text{expected edge degree}}{\text{expected triangle degree}}.$$ 
Hence $\hat{\delta}=n\frac{\text{average edge degree}}{\text{average triangle degree}}$ is a ``good'' estimator of $\delta$. To obtain an estimate of $m$, we first cluster the vertices using spectral clustering on edges and triangles separately, and then compute the respective probability parameters for intra- and inter-cluster connections. Then, we may use $\hat{m} =\frac{\hat{\delta}(\hat{a}_t-\hat{b}_t)}{\hat{a}_e-\hat{b}_e}$ as an estimate of $m$.

The above results also allow us to bound the error rate of spectral clustering of a weighted motif adjacency matrix under the non-uniform hypergraph SBM. Let $A_{W}=A_{E^2} + w A_{T^2}$ be the weighted sum of adjacency matrices of edges and triangle hyperedges with known relative weight $w>0$. Clearly, $E[A_{W}] = E[A_{E^2}] + w E[A_{T^2}]$ and the smallest non-zero eigenvalue of $E(A_{W})$ is $\lambda_{\min}(E[A_{W}]) = (a_e-b_e) + w(n-2)(a_t-b_t)$.    Then, with probability at least $1-o(1)$ we have
\[
\|A_W - E[A_W]\|_2 \leq \| A_{E^2}-E[A_{E^2}]\|_2 + w \|A_{T^2}-E[A_{T^2}]\|_2 \lesssim \sqrt{\Delta} + w \sqrt{\Delta_{t}},
\]
and the error rate is upper bounded according to,
\[
R_{W}  \lesssim \left(\frac{ \sqrt{\Delta} + w \sqrt{\Delta_t}}{ ( a_e-b_e) + w n(a_t-b_t)}\right)^2 \asymp \left(\frac{ \sqrt{a_e} + w \sqrt{na_t}}{ ( a_e-b_e) + w n(a_t-b_t)}\right)^2.
\]
When the asymptotic relationships of Equation (\ref{NUHsetup}) hold, we can further simplify this expression to
\begin{equation}
  R_{W}  \lesssim \left(\frac{1+\sqrt{\frac{n}{\delta}}w}{1+\frac{mn}{\delta}w}\right)^2\frac{a_e}{(a_e-b_e)^2}.
\label{weightnonunif}  
\end{equation}
While Theorem \ref{tradeoff} suggests that depending upon the values of $\delta, m, n$, either the edge-based or triangle-based adjacency matrix has a lower error rate, in practice it might be beneficial for numerical stability to use a weighted average of both of them. The result in Equation \ref{weightnonunif} provides a bound for any weighted sum of these two hyperedge adjacency matrices.

\subsection{The classical SBM}
For the case of a classical SBM, we are only presented with $G_e$ but not $G_t$. In this case, $A_E$ is the adjacency matrix of the graph $G_e$, which we denoted by $A_{E^2}$. The matrix $A_T$ is the triangle motif adjacency matrix constructed from the triangles that arise due to $G_e$, which we denoted by $A_{E^3}$. The smallest non-zero eigenvalue $\lambda_{\min}$ of $E[A_{E^2}]$ equals $\frac{n}{k}\frac{a_e-b_e}{n} = \frac{a_e-b_e}{k}$. In Equation ~(\ref{lambdaPTE}) we described the smallest non-zero eigenvalue for $E[A_{E^3}]$. 

Let $\hat{C}^{(E^2)}$ and $\hat{C}^{(E^3)}$ denote the $n \times k$ matrices of eigenvectors corresponding to the $k$ largest in absolute value eigenvalues of the classical edge-based adjacency matrix and the triangle-adjacency matrix, respectively.
 Using the bound for $\|A_{E^3}-E[A_{E^3}]\|_2$ from Theorem~\ref{ATE}, Section 3.1.3, and the Davis-Kahan Theorem we have the following result.   
  \begin{thm}
 Let $G_e$ be a dyadic graph generated from the $k$-block balanced SBM with parameters $C,n,k,a_e,b_e$.  Then, with probability at least $1-n^{-c}$, the misclustering rate of the community assignments obtained by higher-order spectral clustering equals
\[
R_{E^3} \leq 64(2+\epsilon)\frac{\|A_{E^3}-E[A_{E^3}]\|_2^2}{(\lambda_{\min}(E[A_{E^3}]))^2} \lesssim \frac{64(2+\epsilon)k^4n^2D_{E^3}}{(kb_e^2 + a_e^2+a_eb_e-2b_e^2)^2(a_e-b_e)^2}.
\]
\label{REEE}
 \end{thm}

For the case $k=2$ which is a widely analyzed setting in the SBM literature, one can simplify the bound above. First, note that in this case $\lambda_{\min}(E[A_{E^3}])$ simplifies to $\frac{a(a-b)(a+b)}{n} - O(\frac{1}{n^2})$, while $D_{E^3}=\max\{\frac{a^5}{n^2},c(\log n)^3 \}$. Furthermore, since $a_e \asymp b_e$, we have $a_e \asymp a_e + b_e$. Thus,
\begin{equation}
R_{E^3}  \lesssim  \frac{n^2\max\{(a_e^5/n^2),
(\log n)^3\}}{a_e^2(a_e-b_e)^2(a_e+b_e)^2} \asymp  \frac{\max\{a_e,
(n^2/a_e^4)(\log n)^3\}}{(a_e-b_e)^2}.
\label{RSBM}
\end{equation}

We conclude this section by evaluating the performance of spectral clustering on the weighted sum of the two motif adjacency matrices, the edge-based matrix $A_{E^2}$ and the triangle-based matrix $A_{E^3}$. For this purpose, let $A_{W}=A_{E^2} + w A_{E^3}$ be the weighted sum of motif adjacency matrices, where $w>0$ is a known weight. Clearly, $E(A_{W}) = E[A_{E^2}] + w E[A_{E^3}]$. Then, from the results of Theorem~\ref{ATE} and Theorem 5.2 of~\cite{lei2015consistency}, we have that with probability at least $1-o(1)$, it holds that
\[
\|A_W - E[A_W]\|_2 \leq \| A_{E^2}-E[A_{E^2}]\|_2 + w \|A_{E^3}-E[A_{E^3}]\|_2 \lesssim \sqrt{\Delta} + w \sqrt{D_{E^3}}.
\]
The smallest non-zero eigenvalue of $E[A_W]$ can be computed as
\[
\lambda_{\min}(E[A_W])=( a_e-b_e) + w\frac{(a_e(a_e+b_e) (a_e-b_e)}{n} - O (1/n^2) .
\]
From Equation~(\ref{misclus}), we have
\[
R_{W} \lesssim \left(\frac{ \sqrt{\Delta} + w  \sqrt{D_{E^3}}}{ (a_e-b_e) + w  \frac{(a_e(a_e+b_e) (a_e-b_e)}{n}}\right)^2.
\]

\subsection{Remarks}

Let us start by comparing the upper bound $R_{E^3}$ for higher-order spectral clustering under the SBM obtained in Equation (\ref{RSBM}) with the corresponding upper bound for spectral clustering based on the edge-based adjacency matrix $A_{E^2}$, which reads as
$R \lesssim \frac{a_e}{(a_e-b_e)^2}$ \citep{lei2015consistency}. The bound based on the triangle motif adjacency matrix $A_{E^3}$ is essentially equal to the bound based on the edge adjacency matrix as long as $a_e \gtrsim n^{2/5 + \epsilon}$, or equivalently, as long as $p^e_{\max} \gtrsim \frac{n^{2/5 + \epsilon}}{n}$. However, when $a_e$ grows slower than this rate, the performance guarantees for spectral clustering based on the motif adjacency matrix is worse than the corresponding bound based on the edge adjacency matrix. This result is intuitively justified as we expect very few triangles in a sparse dyadic graph. The presence of a triangle is a random phenomenon rather than an indicator of community structure. Hence, using triangles for community detection could lead to unwanted errors unless the graph is dense. For $a_e \gtrsim n^{2/5 + \epsilon}$, we say that the graph is ``triangle-dense'' and in this case one can use the triangle-adjacency matrix for community detection. When this condition is not satisfied, we either need to perform some form of regularization or completely dispose of the triangle based adjacency matrix.

However, as previously observed, real world networks contain more triangles and higher-order structures, and consequently have a higher level of local clustering than one would expect from the SBM. Hence, the SupSBM is a more appropriate model for networks with community structures. The upper bound on the misclustering rate in Equation~(\ref{Rsup}) suggests that spectral clustering based on higher-order structures can consistently detect communities under the SupSBM. In fact, if $a_e \asymp a_t$, then the underlying upper bound is smaller by a factor of $n$ compared to that of the spectral clustering under the standard SBM. This suggests that even though spectral clustering based on higher-order structures may not be appropriate for the SBM, it offers improved performance for the SupSBM. Note that as we focus on higher-order structures based spectral clustering, we did not analyze edge-adjacency matrix based spectral clustering under the SupSBM. We hence cannot compare the misclustering rate of spectral clustering on the triangle-adjacency matrix with that of the edge-adjacency matrix under SupSBM, as the later does not follow directly from the existing results, e.g.,~\cite{lei2015consistency} (due to the fact that the observed edge-adjacency matrix has edges generated by triangles from $G_t$ in addition to edges from $G_e$). Nevertheless, our analysis of the non-uniform hypergraph SBM, especially Theorem~\ref{tradeoff}, describes the error rate tradeoff between spectral clustering with edge-based and triangle-based adjacency matrices. 

\section{Experiments on Real Data} \label{sec:data}

We test the effectiveness of spectral clustering using a weighted sum of adjacency and Laplacian matrices for higher-order structures on three benchmark network datasets. In particular, we choose to work with a uniformly weighted edge-triangle adjacency matrix, $A_W=A_E+A_T$, where $A_E$ and $A_T$ are the observed edge and triangle adjacency matrices defined earlier. The normalized Laplacian matrix is obtained as $L_w=D_w^{-1/2}A_WD_W^{-1/2}$, where $D_W$ is a diagonal matrix such that $(D_W)_{ii}=\sum_{j}(A_{W})_{ij}$. We compare the performance of various known forms of spectral clustering methods based on edge-based matrices, namely those using adjacency matrices (spA), normalized Laplacian matrices (spL), and regularized normalized Laplacian matrices (rspL)~\citep{sarkar2015role,chin2015stochastic,qr13} with their weighted higher-order structure counterparts, hospA, hospL and horspL, respectively. In all six instances of the spectral clustering, the eigenvectors are row-normalized before applying the k-means algorithm. Table \ref{tab:polblogs} summarizes the performance of the methods.

\textbf{Political blogs data.} The political blog datasets \citep{adamic05}, collected during the 2004 US presidential election, comprise $1490$ political blogs with hyperlinks between them, giving rise to directed edges. These benchmark datasets have been analyzed by a number of authors~\citep{kn11,amini13,qr13,joseph2016impact,j15,gao2017achieving,pc16} in order to test community detection algorithms. Following previous approaches, we first convert directed edges into undirected edges by assigning an edge between two vertices if there is an edge between them in either direction and consider the largest connected component of the resultant graph which contains $1222$ vertices. The ground truth community assignment used for comparisons splits the graph in two groups, liberal and conservative, according to the political leanings of the blogs.  We note the hospA and horspL are competitive with the corresponding edge based methods spA and rspL, respectively. However, for spectral clustering based on the normalized Laplacian matrix, the edge-based method spL completely fails to detect the community structure due to well-documented reasons described in~\cite{qr13,j15,joseph2016impact,gao2017achieving}. On the other hand, hospL succeeds in splitting the graph into two communities with only $59$ misclustered vertices.

\begin{table}
\protect\caption{The number of misclustered vertices for various spectral community detection algorithms that use different forms of weighted higher-order matrices. Performance is evaluated based on a known ground truth model.}

\centering
\begin{tabular}{ccccccccc}
\hline
Dataset & spA & hospA & spL & hospL & rspL & horspL \tabularnewline
\hline
Political blogs & 63  & 71 & 588 & 59 & 64 & 64 \tabularnewline
Karate club & 0 & 0 & 1 & 0 & 0 & 0 \tabularnewline
Dolphins & 2 & 2 & 2 & 1 & 2 & 1  \tabularnewline
\hline
\end{tabular}
\label{tab:polblogs}
\end{table}
\textbf{Karate club data.} The Zachary's karate club data \citep{zachary77} is another frequently used benchmark dataset for network community detection \citep{ng04,bc09,j15}. The network describes friendship patterns of $34$ members of a karate club and the ground truth splits club members into two subgroups. The method spL misclusters one vertex, while all other methods manage to recover the communities in an error-free manner.

 \textbf{Dolphin social network data.} This dataset describes an undirected social network involving $62$ dolphins in Doubtful Sound, New Zealand, curated by~\citet{dolphin2}. Over the course of the study the group split into two due to departure of a ``well connected'' dolphin. These two subgroups are used as the ground truth. From Table~\ref{tab:polblogs}, one can see that only hospL and horspL miscluster one dolphin, while all the remaining methods miscluster two dolphins.

\section{Conclusion and future directions}
We proposed and analyzed a superimposed stochastic block model, a mathematically tractable random graph model with community structure, that produces networks with properties similar to that observed in real networks. In particular it can generate sparse networks with short average path length (small-world), strong local clustering, and community structure. To produce the strong local clustering property, the model allows for dependencies among the edges, yet remaining mathematically suitable for analysis of algorithms. While not pursued here, a degree correction to the model similar to that of degree corrected stochastic block model is expected to produce networks with highly heterogeneous degree distribution (power law), hub nodes and core-periphery structure while simultaneously retaining the aforementioned properties. We hope to extend the model in that direction in a future work. 

We have also analyzed the performance of the higher-order spectral clustering algorithm under the proposed SupSBM. This analysis showed that it is possible to mathematically analyze community detection algorithms under the supSBM, and that the method can detect community structure consistently for graphs generated from the SupSBM. In future, we hope to determine minimax rates of error of community detection under the SupSBM and obtain algorithms that achieve those rates.

\section*{Appendix A}
In the Appendices we will use $r$ and $c$ to represent generic constants whose values will be different for different results.
\subsubsection*{Proof of Theorem \ref{ATT}}
\begin{proof}
We follow and extend the arguments in the proof of a similar result for standard adjacency matrices in~\citet{lei2015consistency,gao2017achieving}, and \citet{chin2015stochastic} to the case of triangle-motif adjacency matrices. The arguments in all of the above mentioned papers rely on the use of $\epsilon-$nets on random regular graphs~\citep{friedman1989second,feige2005spectral}. 

Let $S$ denote the unit sphere in the $n$ dimensional Euclidean space. An $\epsilon-$net of the sphere is defined as follows:
\[
\mathcal{N}=\{x=(x_1,\ldots,x_n) \in S: \, \forall i, \, \epsilon\sqrt{n}x_i \in \mathbb{Z}\}, 
\]
where $\mathbb{Z}$ denotes the set of integers. Hence, $\mathcal{N}$ is a set of grid points of size $\frac{1}{\epsilon \sqrt{n}}$ spanning all directions within the unit sphere. For our analysis we only use $\epsilon=1/2-$nets of spheres and henceforth use $\mathcal{N}$ to denote such nets.

Next, we recall Lemma 2.1 of~\cite{lei2015consistency} which established that for any $W \in \mathbb{R}^{n\times n}$, one has $\|W\|_2 \leq 4 \sup_{x,y \in \mathcal{N}} |x^TWy|$. Hence, a constant-approximation upper bound for $\|A_{T^2}-E[A_{T^2}]\|_2$ may be found by optimizing $|x^T(A_{T^2}-E[A_{T^2}]) y|$ over all possible pairs $(x,y) \in \mathcal{N}$. In addition, note that
\begin{equation}
x^T(A_{T^2}-E[A_{T^2}]) y =\sum_{i,j} x_iy_j(A_{T^2}-E[A_{T^2}])_{ij}=\sum_{i,j}\sum_{k \neq i,j}x_iy_j(T_{ijk}-E[T_{ijk}]).
\end{equation}
We now divide the pairs $(x_i,y_j)$ into two sets, the set of \textit{light pairs} $L$ and the set of \textit{heavy pairs} $H$, according to
\begin{align*}
L & =\{(i,j): |x_iy_j| \leq \frac{\sqrt{\Delta_t}}{n}\}, \\
H & =\{(i,j): |x_iy_j| > \frac{\sqrt{\Delta_t}}{n}\},
\end{align*}
where $\Delta_t$ is as defined in the statement of the theorem.

We bound the term $x^T(A_{T^2}-E[A_{T^2}]) y$ separately for the light and heavy pairs, as summarized in the following two lemmas. 

\begin{lem}
(\textit{Light pairs}) For some constant $r_1>0$, there exists a constant $c_2(r_1)>0$, such that with probability at least $1-\exp(-r_1n)$,
\[
\sup_{x,y \in T} |\sum_{(i,j) \in L} \sum_{k} x_{i}y_{j}(T_{ijk}-E[T_{ijk}])| < c_2(r_2)\sqrt{\Delta_t}.
\]
\label{lightpairs}
\end{lem}
Whenever clear from the context, we suppress the dependence of the constants on other terms (e.g., $c_2(r_2)=c_2$.)

To obtain a similar bound for heavy pairs, we first note that
\begin{equation}
\sup_{x,y \in T} |\sum_{(i,j) \in H} \sum_{k} x_{i}y_{j}w_{ijk}| \leq \sup_{x,y \in T} |\sum_{(i,j) \in H} \sum_{k} x_{i}y_{j}a_{ijk}| + \sup_{x,y \in T} |\sum_{(i,j) \in H} \sum_{k} x_{i}y_{j}p_{ijk}|.
\label{heavy pairs}
\end{equation}
The second term can be easily bounded as follows:
\begin{align*}
|\sum_{(i,j) \in H} \sum_{k} x_{i}y_{j}p_{ijk}| & \leq \sum_{(i,j) \in H} \sum_{k} \frac{x_{i}^2y_{j}^2}{|x_{i}y_{j}|}p_{ijk} \\
& \leq \frac{n}{\sqrt{\Delta_t}}\sum_k \max_{i,j,k} (p_{ijk}) \sum_{i,j} x_i^2 y_j^2 \\
& \leq \frac{n}{\sqrt{\Delta_t}} \frac{\Delta_t}{n} \leq \sqrt{\Delta_t}.
\end{align*}
How to bound the first term is described in the next Lemma \ref{heavypairs}.

\begin{lem} For some constant $r_2>0$, there exists a constant $c_3(r_2)>0$ such that with probability at least $1-n^{-r_2}$, $\sum_{(i,j) \in H} \sum_{k} x_{i}y_{j}T_{ijk} \leq c_3 \sqrt{\Delta_t}$.
\label{heavypairs}
\end{lem}

Combining the results for the light and heavy pairs, we find that with probability at least $1-n^{-r}$,
\[
\|A_{T^2}-E[A_{T^2}]\|_2 \leq 4 \sup_{x,y \in T}|x^T(A_{T^2}-E[A_{T^2}])y| \leq c_1 \sqrt{\Delta_t}. 
\]
This completes the proof of Theorem 1.
\end{proof}

\subsubsection*{Proof of Theorem \ref{ATE}}

\begin{proof}
As before, we create an $\epsilon-$ net $\mathcal{N}$ for the unit sphere and separately analyze the light and heavy pairs. In this setting, the pairs are defined according to
$$L  =\{(i,j): |x_iy_j| \leq \frac{\sqrt{D_{E^3}}}{n\tau_{\max}}\}$$ 
and 
$$H  =\{(i,j): |x_iy_j| > \frac{\sqrt{D_{E^3}}}{n\tau_{\max}}\},$$
with $D_{E^3}$ as defined in the statement of the Theorem.

For the light pairs, we can prove the following result.
\begin{lem}
(\textit{Light pairs}) For some constant $r_1>0$, there exists a constant $c_3(r_1)>0$ such that with probability at least $1-n^{-r_1}$,
\[
\sup_{x,y \in \mathcal{N}} |\sum_{(i,j) \in L} x_{i}y_{j}(A_{E^3} -E[A_{E^3}])_{ij}| < c_3\sqrt{D_{E^3}}.
\]
\label{TElight}
\end{lem}

To bound the contribution of the heavy pairs, we once again divide the sum into two terms. 

First, let $W_{E^3} = (A_{E^3} -E[A_{E^3}])$ and note that $\max_{i,j} (E[A_{E^3}])_{ij} \leq n(p^e_{\max})^3$. Then, 
\begin{equation}
\sup_{x,y \in \mathcal{N}} |\sum_{(i,j) \in H} x_{i}y_{j}(W_{E^3})_{ij}| \leq \sup_{x,y \in \mathcal{N}} |\sum_{(i,j) \in H}  x_{i}y_{j}(A_{E^3})_{ij}| + \sup_{x,y \in \mathcal{N}} |\sum_{(i,j) \in H}  x_{i}y_{j}(E[A_{E^3}])_{ij}|.
\label{heavy pairs TE}
\end{equation}
The second term can be bounded as follows:
\begin{align*}
|\sum_{(i,j) \in H}  x_{i}y_{j}(E[A_{E^3}])_{ij}| & \leq \sum_{(i,j) \in H}  \frac{x_{i}^2y_{j}^2}{|x_{i}y_{j}|}(E[A_{E^3}])_{ij} \\
& \leq \frac{n\tau_{\max}}{\sqrt{D_{E^3}}}\max_{ij} (E[A_{E^3}])_{ij}\sum_{(i,j)} x_i^2 y_j^2 \\
& \leq \frac{n\tau_{\max}}{\sqrt{D_{E^3}}}n(p^e_{\max})^3
\leq \frac{D_{E^3}}{\sqrt{D_{E^3}}} \leq \sqrt{D_{E^3}},
\end{align*}
where the penultimate inequality follows since if $\tau_{\max}=n(p^e_{\max})^2$, then 
$$n\tau_{\max}n(p^e_{\max})^3 = n^3(p^e_{\max})^5 \leq D_{E^3}.$$ In addition, if $\tau_{\max}=\log n$, then $n(p^e_{\max})^2 \leq \log n$. Consequently, 
$$n\tau_{\max}n(p^e_{\max})^3 = n(p^e_{\max})n(p^e_{\max})^2 \log n \leq n(p^e_{\max}) (\log n)^2 \leq D_{E^3}.$$

For the first term in Equation (\ref{heavy pairs TE}) we have the following result.
\begin{lem}  For some constant $c>0$, there exists a constant $c_1(c)>0$ such that with probability at least $1-2n^{-c}$, $\sum_{(i,j) \in H} \sum_{k} x_{i}y_{j}(A_{E^3})_{ij} \leq c_1 \sqrt{D_{E^3}}$.
\label{TEheavy1}
\end{lem}

Combining the results for the light and heavy pairs, we obtain
\[
\|A_{E^3}-E[A_{E^3}\| \leq 4 \sup_{x,y \in \mathcal{N}}|x^T(A_{E^3}-E[A_{E^3}])y| \leq c_1 \sqrt{D_{E^3}}. 
\]
\end{proof}

\subsubsection*{Proof of Theorem \ref{ATTT}}
\begin{proof}

The proof of this result and those of Theorems \ref{ATTE} and \ref{ATEE} will repeatedly use Theorem 9 of \cite{warnke2017upper}, which we reproduce here for ease of reference.

\begin{prop}
[Theorem 9 of \cite{warnke2017upper}] Let $(Y_{i}),\, i \in \mathcal{I}$ be a collection of non-negative random variables with 
$\sum_{i \in \mathcal{I}} E(Y_{i}) \leq \mu$. Assume that $\sim$
is a symmetric relation on $\mathcal{I}$ such that each $Y_{i}$ with $i \in \mathcal{I}$ is independent of $\{Y_{j}:\, j \in \mathcal{I}, \, j \nsim  i \}$. Let
$Z_C = \max \sum_{i \in \mathcal{J}} Y_{i}$, where the maximum is taken over all sets $\mathcal{J} \subset \mathcal{I}$ such that $\max_{j \in \mathcal{J}} \sum_{i \in \mathcal{J}, i \sim j} Y_{i} \leq C$. Then for all $C, t > 0$ we have
\begin{align*}
P(Z_C \geq \mu + t ) \leq \min \Big\{ \exp \left( -\frac{t^2}{2C(\mu +t/3)}\right), \left(1+\frac{t}{2\mu}\right)^{-t/2C} \Big \}.  
\end{align*}
\label{prop1}
\end{prop}

For any vertex $i$, define the degree of $i$ in the matrix $A_{T^3}$ according to
\[
(d_{T^3})_i = \sum_{j}\sum_{k} T^3_{ijk}.
\]
 
The expectation of the degree may be bounded as  
\begin{align*}
E[(d_{T^3})_i] & =E[\sum_{j} \sum_{k}(1-T_{ijk}) 1(\sum_{k_1 \neq k}T_{ijk_1}>0)1(\sum_{k_2 \neq k}T_{jkk_2}>0)1(\sum_{k_3\neq k}T_{ikk_3}>0)]\\
& \leq  \sum_{j} \sum_{k} P(\sum_{k_1 \neq k}T_{ijk_1}>0)P(\sum_{k_2 \neq k}T_{jkk_2}>0)P(\sum_{k_3\neq k}T_{ikk_3}>0) \\
& \leq  \sum_{j} \sum_{k} (np_{\max}^{t})^3 \leq n^5(p^{t}_{\max})^3 \leq  \Delta_{T^3},
\end{align*}
where the second inequality follows since 
\[
P(\sum_{k_1 \neq k}T_{ijk_1}>0) \leq P(\cup_{k_1 \neq k}\{T_{ijk_1}=1\}) \leq \cup_{k_1 \neq k} P(\{T_{ijk_1}=1\}) \leq np^t_{\max}.
\]

Let $I_{i}=\{(T^3)_{ijk}, j=\{1,\ldots,n\},k=\{1,\ldots,n\}\}$ denote the set of all triangles incident to vertex $i$ and generated incidentally by three other triangles in $G_t$. Observe that the set $\{T^{3}\}$ comprises elements that are indicator random variables indexed by $\theta=\{i,j,k\}$ corresponding to incidentally generated triangle. Consequently, two random variables in the family $(T^3)_{\theta}$, when restricted to the set $I_i$, are dependent if and only if one of the triangles creating the edges $ik$ or $ij$ of $(T^3)_{ijk}$ includes $j'$ or $k'$ as a vertex and is consequently part of $(T^3)_{ij'k'}$ (see Figure~\ref{dependence}(b)). We refer to an event corresponding to the above described scenario as $TC$ and note that this event also accounts for the case when we have $T^3_{ij'k}$ ``sharing'' the edge $ik$ with $T^3_{ijk}$.

In summary, the set $I_i$ contains $O(n^2)$ dependent random variables, with $(d_{T^3})_i$ denoting the sum of all the random variables in the set $I_i$. However, in what follows we show that the dependence is ``limited" in the sense that we can limit the number of other variables dependent on one random variable in the set $I_i$ to $O(n^4(p^t_{\max})^3)$ with high probability. 

We characterize the event $TC$ through an associated indicator random variable. We show that the number of incidentally generated triangles $T^3_{ij'k'}$ that give rise to TC events is bounded, provided that certain ``good events" occur with high probability. For this purpose, let $T_{ij'k}$ be a triangle in $G_t$ leading to the creation of an incidental triangle $T^3_{ijk}$ (see Figure~\ref{dependence}(b)). To create the incidental triangle $T^3_{ij'k'}$, we also require the existence of at least two triangles from $G_t$ with edges $ik'$ and $jk'$. We capture this event through its indicator variable 
\[
V_{j'k'} = (1-T_{ij'k'})T_{ij'k}1(\sum_{k''\neq i}T_{j'k'k''}>0)1(\sum_{k'''}T_{ik'k'''}>0).
\]
Observe that one can think of $V_{j'k'}$, as the random variable $T^3_{ij'k'}$ given that $T^3_{ijk}=1$ (see Figure~\ref{dependence}(b)). Consequently, for any $T^3_{ijk} \in I_i$, the number of incidentally generated triangles $T^3_{ij'k'}$ that also belong to the set $I_i$ \textit{and} contribute to the occurrence of the event $TC$ is at most $2 \sum_{j'}(\sum_{k'} V_{j'k'})$.

Next, we define a ``good event" as $\Gamma = \Gamma_1 \cap \Gamma_2$, where $\Gamma_1$ and $\Gamma_2$ are two events that for any $i,j,k$ may be described as follows: 
\begin{align*}
\Gamma_1 & = \{\text{For an edge } ij\text{ there are at most } V_{\max}= \max \{n^3(p^t_{\max})^2,(\log n)^2\} \text{ vertices } k' \\
& \quad \text{ such that the edges } ik' \text{ and } jk' \text{ are introduced by triangles from } G_t\}, \\
\Gamma_2 & = \{ \text{The number of triangles in } G_t \text{ sharing an edge } ij \text{ is at most } 3W_{\max}=3\max\{np^t_{\max}, \log n \} \}.
\end{align*}

Hence, the event $\Gamma_2$ essentially asserts that there are $6W_{\max}$ choices for the value of $j'$, and given a $j'$, the event $\Gamma_1$ asserts that there are $V_{\max}$ choices for a $k'$. Consequently, under the ``good event" $\Gamma$ the number of triangles in $I_i$ on which a triangle $T^3_{ijk}$ depends on is $2\sum_{j'}\sum_{k'}V_{j'k'} \leq 6V_{\max}W_{\max}$.  

Next, define a set $J \subset I_i$ as follows:
\[
J=\{ \theta \in I_i : \, \max_{\theta_1 \in J} |\theta_2 \in J;\theta_1 \text{ and } \theta_2 \text{ are dependent } | \leq 6V_{\max}W_{\max} \}.
\]

In a nutshell, the sets $J$ are collections of $\theta$'s such that no $T^3_{\theta_1}$ is dependent on more than $6V_{\max}W_{\max}$ other $T^3_{\theta_2}$'s. Let $\theta_1 \sim \theta_2$ state that the two underlying random variables indexed by $\theta_1$ and $\theta_2$ are dependent. Then we have,
\[
\max_{\theta_2 \in J} \sum_{\theta_1 \in J: \theta_1 \sim \theta_2} T^3_{\theta_1} =  2\sum_{j'}\sum_{k'}V_{j'k'}\leq 6V_{\max}W_{\max},   \quad E [ \sum_{\theta \in I_i}T^3_{\theta}] = E[(d_{T^3})_i] \leq  \Delta_{T^3}. 
\]
Applying Proposition \ref{prop1} for $t=\mu=\Delta_{T^3}$ leads to
\begin{align*}
    P(\max_{J} \sum_{\theta \in J}T^3_{\theta} \geq 2\Delta_{T^3})  
    & \leq \min \bigg \{\exp \left( -\frac{\Delta_{T^3}^2}{12V_{\max}W_{\max}(\Delta_{T^3} + \Delta_{T^3}/3)}\right), \left(1+\frac{\Delta_{T^3}}{2\Delta_{T^3}}\right)^{-\frac{\Delta_{T^3}}{12V_{\max}W_{\max}}} \bigg \}\\
    & = \min \bigg \{\exp \left(- \frac{\Delta_{T^3}}{ \frac{48}{3}V_{\max}W_{\max}}\right),\frac{3}{2}^{-\Delta_{T^3}/12V_{\max}W_{\max}} \} \\
    & \leq \exp(-c' \log n) \leq n^{-c'}.
\end{align*}

The last inequality may be established through the following argument. If $W_{\max} =np^t_{\max}$, then $np^t_{\max} \geq \log n$, which implies
$$n^3(p^t_{\max})^2 \geq n^3 (\frac{\log n}{n})^2 = n (\log n)^2.$$ 
Then, $V_{\max}=n^3(p^t_{\max})^2$, and consequently 
$$\frac{\Delta_{T^3}}{V_{\max}W_{\max}}  = \max \{n, \frac{(\log n)^4}{n^4(p^t_{\max})^3} \} \geq n.$$ 
On the other hand, if $W_{\max} =\log n$, then $np^t_{\max} < \log n$. Now, either $V_{\max}=(\log n)^2$, in which case $W_{\max}V_{\max}=(\log n)^3$ and  $\frac{\Delta_{T^3}}{V_{\max}} \geq \log n$. Or, $V_{\max}=n^3(p^t_{\max})^2$, and consequently $V_{\max}W_{\max}=n^3(p^t_{\max})^2 \log n$. Then 
$$\frac{\Delta_{T^3}}{V_{\max}W_{\max}} = \max \{\frac{n^2p^t_{\max}}{\log n}, \frac{(\log n)^4}{n^3(p^t_{\max})^2 \log n} \}\geq \log n,$$ 
since $n^2p^t_{\max} > (\log n)^2$ by assumption.

Recall that the event $TC$ describes the only setting for which two random variables in the set $I_i$ are dependent on each other;  under the good event $\Gamma$, we have $I_i=\argmax_J |J|$ and consequently, $\max_{J} \sum_{\alpha \in J}T^3_{\alpha}=(d_{T^3})_i$. 

Next, we need to show that the probability of the ``bad event'' (i.e., complement of the good event) is exponentially small. 
For that, we note
\[
P(\Gamma^C) = P(\Gamma_1^C \cup \Gamma_2^C) \leq P(\Gamma_1^C)+P(\Gamma_2^C).
\]
The last term $P(\Gamma_2^C)$ can be easily bounded using Bernstein's inequality as follows. 
Let $W_{ij}=\sum_{k}T_{ijk}$. Then $W_{ij}$ counts the number of triangles in $G_t$ sharing an edge $ij$. The event $\Gamma_2$ asserts that the number of triangles in $G_t$ sharing an edge is at most $3W_{\max}=3\max\{np^t_{\max}, \log n \}$.
From Bernstein's inequality and the union bound we consequently have
\begin{align*}
    P(\Gamma_2^C) & \leq n^2 P (W_{ij} > 3W_{max}) \\
    & \leq n^2 \exp(-\frac{9W^2_{\max} }{2\sum_{k}p^t_{ijk}(1-p^t_{ijk})+\frac{6}{3}W_{\max}}) \\
    & \leq n^2 \exp(-\frac{9W^2_{\max}}{2W_{\max}+2W_{\max}}) \\
    &\leq n^2 \exp(-\frac{9}{4}W_{\max}) \leq \exp(-\frac{1}{4}\log n) \leq n^{-c''}.
\end{align*}

\begin{figure}[h]
\centering
\begin{subfigure}{0.3\linewidth}
\includegraphics[width=\linewidth]{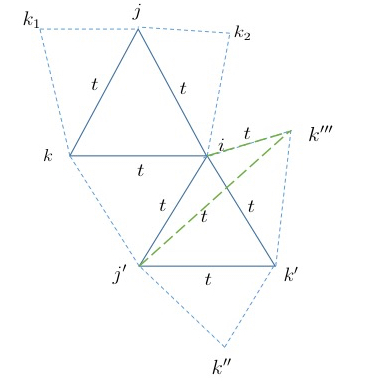}
 \end{subfigure}%
 \hspace{40pt}
 \begin{subfigure}{0.3\linewidth}
\includegraphics[width=\linewidth]{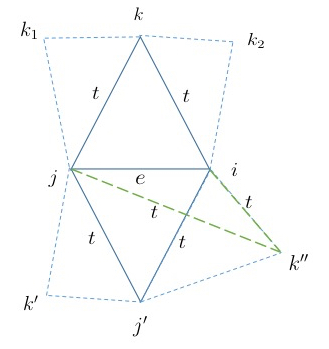}
 \end{subfigure}
   \begin{center}
     (a) \hspace{200pt} (b)
 \end{center}
 \begin{subfigure}{0.3\linewidth}
\includegraphics[width=\linewidth]{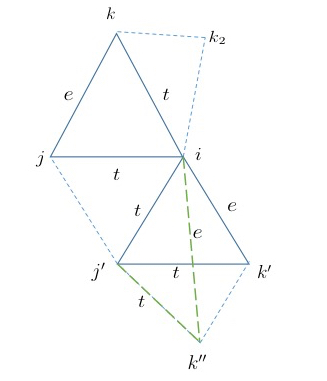}
 \end{subfigure}%
 \hspace{40pt} 
\begin{subfigure}{0.3\linewidth}
\includegraphics[width=\linewidth]{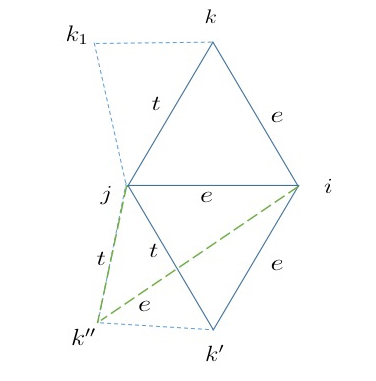}
 \end{subfigure}
  \begin{center}
     (c) \hspace{200pt} (d)
 \end{center}
 \caption{Second-order dependencies that need to be taken into account in the concentration inequalities for ``good events":  (a) $\Gamma_1$ for $T^3$, (b) $\Gamma_1$ for $T^2E$, (c) $\Gamma_3$ for $T^2E$, and (d) $\Gamma_3$ for $TE^2$.}
 \label{secdep}
\end{figure}

We now turn our attention to the event $\Gamma_1$. Fix a $j'$. The sum $\sum_{k'}V_{j'k'}$ includes dependent random variables; two random variables in the sum, say $V_{j'k'}$ and $V_{j'k''}$, are dependent if and only if there is a triangle, say $T_{ik'k''}$ in $G_t$ that has both $ik'$ and $ik''$ as edges (Figure~\ref{secdep}(a)). 
However, the number of triangles in $G_t$ sharing an edge $ij'$ can be bounded by referring to the event $\Gamma_2$. First, we define $I_{ij'}$ to be the collection of all $V_{j'k'}$ with fixed $i$ and $j'$. Then, we define the sets $J$s as subsets of $I_{ij'}$ such that no $V_{j'k'}$ is dependent on more than $W_{\max}$ other $V_{j'k'}$s.  To apply Proposition \ref{prop1} to $\sum_{k'} V_{j'k'}$, we first observe that one may upper bound the relevant expectation as
\[
E[\sum_{k'}(1-T_{ij'k'})1(\sum_{k''\neq i}a^t_{j'k'k''}>0)1(\sum_{k'''}a^t_{ik'k'''}>0)] \leq n^3(p^t_{\max})^2 \leq V_{\max},
\]
and
$\max_{\beta \in J} \sum_{\alpha \in J: \alpha \sim \beta} V_{j'k'} \leq W_{\max}.$
Then, with $t=\mu=V_{\max}$,
\begin{align*}
    P(V_{ij'} \geq 2V_{\max})  
    & \leq \min \bigg \{\exp \left( -\frac{V_{\max}^2}{2W_{\max}(V_{\max} + V_{\max}/3)}\right), \left(1+\frac{V_{\max}}{2V_{\max}}\right)^{-V_{\max}/2W_{\max}} \bigg \}\\
    & = \min \bigg \{\exp \left(- \frac{V_{\max}}{ \frac{8}{3}W_{\max}}\right),\frac{3}{2}^{-V_{\max}/2W_{\max}} \} \\
    & \leq \exp(-c''' \log n) \leq n^{-c'''}.
\end{align*}
The last inequality holds due to the following argument. If $W_{\max} =np^t_{\max}$, then $p^t_{\max} \geq \frac{\log n}{n}$, and consequently, $n^2p^t_{\max} \geq n \log n$. Then $\frac{V_{\max}}{W_{\max}} \geq n^2p^t_{\max}>\log n$. If $W_{\max} =\log n$, then $\frac{V_{\max}}{W_{\max}} \geq \log n,$ since $V_{\max} \geq (\log n)^2$. Since there are at most $n$ choices for $j'$, for any $i$, the union bound leads to 
$$P(\Gamma_1^C)\leq nP(V_{ij'} \geq 2V_{\max}) \leq n^{-c''''}.$$

Combining the results we have
\[
P((d_{T^3})_i \geq 2 \Delta_{T^3}) \leq n^{-c'} +n^{-c''} +n^{-c''''}+ n^{-c''}.
 \]
Invoking the union bound, now over all $i$, we can show that $\max_i (d_{T^3})_i \leq c_1 \Delta_{T^3}$ with probability at least $1-n^{-c''}$. By Equation (\ref{twonorm}), the claimed result holds.
\end{proof}

\subsubsection*{Proof of Theorem \ref{ATTE}}
\begin{proof}
Triangles of type $T^2E$ are generated by two triangles from $G_t$ and one edge from $G_e$. Without loss of generality, we may assume that in $T^2E_{ijk}$, the sides $ij$ and $jk$ are generated by triangles from $G_t$ and that the side $ik$ is generated by an edge from $G_e$. 
Then, we have
\begin{align*}
E[(d_{T^2E})_{i}] & =E[\sum_{j} \sum_{k} T^2E_{ijk}]\\
& \leq  \sum_{j} \sum_{k} P(\sum_{k_1 \neq k}T_{ijk_1}>0)P(\sum_{k_2 \neq k}T_{jkk_2}>0)P(E_{ik}=1) \\
& \leq  \sum_{j} \sum_{k} (np_{\max}^{t})^2(p^e_{\max}) \leq n^4(p^{t}_{\max})^2p^e_{\max} \leq  \Delta_{T^2E}.
\end{align*}
Let the set $I_i=\{(T^2E)_{ijk}, j=\{1,\ldots,n\},k=\{1,\ldots,n\}\}$  denote the set of all incidentally generated triangles of type $T^2E$ that includes the vertex $i$. The set $\{T^2E_{\theta}\}$, indexed by $\theta=\{i,j,k\}$, represents a family of indicator variables corresponding to incidentally generated triangles of type $T^2E$. Two random variables in the family $(T^2E)_{\theta}$ restricted to the set $I_i$ may be dependent in two scenarios. One possibility is that one of the sides $ij$ or $ik$ is an edge from $G_e$ and serves as an edge for  $(T^2E)_{ijj'}$ or $(T^2E)_{ikj'},$ for some $j'$ (see Figure \ref{dependence}(c)). The other possibility is that one of the sides $ij$ or $ik$ is created by a triangle from $G_t$ and the same triangle is involved in creating $(T^2E)_{ij'k'}$ for some $j'$ and $k'$ (see Figure \ref{dependence}(d)). We refer to these two events as $TC_1$ and $TC_2$, respectively.

We now need to derive a bound on $(d_{T^2E})_i$, which equals the sum of the random variables $T^2E_{ijk}$, that holds with high probability. We proceed as in the proof of the previous theorem and describe ``good events" which limit the number of random variables that a  random variable in the sum depends on with high probability.

For this purpose, we characterize the events $TC_1$ and $TC_2$ using indicator variables. First, define the random variables
\[Q_{j'}=(1-T_{ijj'})1(\sum_{k'}T_{jj'k'}>0)1(\sum_{k''}T_{ij'k''}>0).
\] 
Then, for any $T^2E_{ijk}$, the number of other incidentally generated triangles in $I_i$ creating the event $TC_1$ is at most $\sum_{j'} Q_{j'}$ (Figure~\ref{dependence}(c)).

With regards to the event $TC_2$, define the following random variable
\[U_{j'k'}=(1-T_{ij'k'})T_{ijj'}1(\sum_{k''\neq i}T_{j'k'k''}>0)1(E_{ik'}=1).
\] 
Then, for any $T^2E_{ijk}$, the number of additional incidental triangles in $I_i$ that contribute to the event $TC_2$ is at most $2\sum_{j'} \sum_{k'}U_{j'k'}$ (Figure~\ref{dependence}(d)).

As before, define a ``good event" as $\Gamma = \Gamma_1 \cap \Gamma_2 \cap \Gamma_3$, where for any $i,j,k$, $\Gamma_1$, $\Gamma_2$ and $\Gamma_3$ are defined as:
\begin{align*}
\Gamma_1 & = \{\text{For an edge } ij\text{ there are at most } V_{\max}= \max \{n^3(p^t_{\max})^2,(\log n)^2\} \text{ vertices } k', \\
& \quad \text{ such that edges } ik' \text{ and } jk' \text{ are generated by triangles from } G_t\}, \\
\Gamma_2 & = \{ \text{The number of triangles in } G_t \text{ incident to an edge } ij \text{ is at most } 3W_{\max}=3\max\{np^t_{\max}, \log n \} \},\\
\Gamma_3 & = \{\text{For an edge } ij\text{ there are at most } U_{\max}= \max \{n^2p^t_{\max}p^e_{\max},(\log n)^2\} \text{ vertices } k', \\
& \quad \text{ such that the edge } ik' \text{ arises from  } G_e \text{ and edge } jk' \text{ arises from a triangle in } G_t\}.
\end{align*}
Note the second event $\Gamma_2$ is the same as the event $\Gamma_2$ described in the proof of Theorem~\ref{ATTT}.

As in the previous setting, the events above are defined in a way that ensures that ``good events" happen with high probability. We note that under the events $\Gamma_1$ and $\Gamma_2$, one has
\[
\sum_{j'}Q_{j'} \leq V_{\max}, \quad 2\sum_{j'}\sum_{k'} U_{j'k'} \leq 6W_{\max}U_{\max}.
\]
Hence, the two events limit the number of occurrences of the events $TC_1$ and $TC_2$, respectively, and consequently limit the number of random variables that a random variable in the sum $(d_{T^2E})_i$ is dependent on. We once again apply Proposition \ref{prop1} to $(d_{T^2E})_i$ under the good event $\Gamma$ to obtain an upper bound for $P(\Gamma^C)$. For this purpose, define a set $J \subset I_i$ as follows:
\[
J=\{ \theta \in I_i : \, \max_{\theta_1 \in J} |\theta_2 \in J;\theta_1 \text{ and } \theta_2 \text{ are dependent } | \leq C \},
\]
where $C$ may be found from
\begin{align*}
\max_{\beta \in J} \sum_{\theta \in J: \theta \sim \beta} T^2E_{\theta} & = \sum_{j'}Q_{j'} + \sum_{j'}\sum_{k'} U_{j'k'} \\
& \leq V_{\max} + 6W_{\max}U_{\max} \\ &\leq 7\max \{n^3(p^t_{\max})^2, n^2p^t_{\max}p^e_{\max}\log n, (\log n )^3  \} =C.
\end{align*}
 Then, $E[\max_{J}\sum_{\alpha \in J}T^2E_{\alpha}] \leq  \Delta_{T^2E}$, and with $t=\mu=\Delta_{T^2E}$,
\begin{align*}
    P( \max_{J}\sum_{\theta\in J}T^2E_{\theta} \geq 2\Delta_{T^2E})  
    & \leq \min \bigg \{\exp \left( -\frac{\Delta_{T^2E}^2}{2C(\Delta_{T^2E} + \Delta_{T^2E}/3)}\right), \left(1+\frac{\Delta_{T^2E}}{2\Delta_{T^2E}}\right)^{-2\Delta_{T^2E}/2C} \bigg \}\\
    & = \min \bigg \{\exp \left(- \frac{\Delta_{T^2E}}{ \frac{8}{3}C}\right),2^{-\Delta_{T^2E}/C} \} \\
    & \leq \exp(-c' \log n) \leq n^{-c'},
\end{align*}
where the last inequality holds due to the following argument. If $C =n^3(p^t_{\max})^2$, then $\frac{\Delta_{T^2E}}{C} \geq np^e_{\max}$ which, by assumption, is greater than $ \log n$. If $C =n^2p^t_{\max}p^e_{\max} \log n$, then $\frac{\Delta_{T^2E}}{C} \geq \frac{n^2p^t_{\max}}{\log n}$ which, by assumption, is greater than $\log n$. Finally, if $C=(\log n )^3,$ then $\frac{\Delta_{T^2E}}{C} \geq \log n$.

In our previous proofs, we already established upper bounds for $P(\Gamma_1^C)$ and $P(\Gamma_2^C)$. To complete the proof of the claimed result, we only need to determine an upper bound on $P(\Gamma_3^C)$. Using the previously introduced variables $U_{j'k'}$, the event $\Gamma_3$ occurs if $ \sum_{k'}U_{j'k'} \leq U_{\max}$ for any $j'$. Note that the sum $\sum_{k'}U_{j'k'}$ includes dependent random variables.

An upper bound on the expectation of this sum reads as
\[
E(\sum_{k'}U_{j'k'}) \leq E[ \sum_{k'}1(\sum_{k''\neq i}T_{jk'k''}>0)1(E_{ik'}=1)] \leq n^2p^t_{\max}p^e_{\max} \leq U_{\max}.
\]
We also introduce the set $J$ that restricts the number of $U$-variables that another $U$-variable in the sum $ \sum_{k'}U_{j'k'}$ is dependent on: 
\[
J=\{ \theta : \, \max_{\theta_1 \in J} |\theta_2 \in J;\theta_1 \text{ and } \theta_2 \text{ are dependent } | \leq W_{\max} \}.
\]
Fix $i$ and $j'$ and define $I_{ij'}$ to be the collection of all random variables $U_{j'k'}$ that contribute to the event $TC_2$. Two random variables in the sum $\sum_{k'}U_{j'k'}$, say $U_{j'k'}$ and $U_{j'k''}$, are dependent (conditioned on $TE^2_{ijk}$) if and only if the triangle $T_{jk'k''}$ from $G_t$ generates an edge for both the incidental triangles characterized by $U_{j'k'}$ and $U_{j'k''}$ (see Figure~\ref{secdep}(c)). The set $\Gamma_3$ essentially limits the frequency of such triangles $T_{jk'k''}$s: under the event $\Gamma_3$, the largest set $J$ as defined above is equal to the set $I_{ij'}$.
We therefore have 
$\max_{\beta \in J} \sum_{\alpha \in J: \alpha \sim \beta} U_{\alpha} \leq W_{\max}$, 
and for $t=\mu=U_{\max}$,
\begin{align*}
    P(\max \sum_{\theta \in J}U_{\theta}\geq 2U_{\max})  
    & \leq \min \bigg \{\exp \left( -\frac{U_{\max}^2}{2W_{\max}(U_{\max} + U_{\max}/3)}\right), \left(1+\frac{U_{\max}}{2U_{\max}}\right)^{-U_{\max}/2W_{\max}} \bigg \}\\
    & = \min \bigg \{\exp \left(- \frac{K_{\max}}{ \frac{8}{3}W_{\max}}\right),\frac{3}{2}^{-U_{\max}/2W_{\max}} \} \\
    & \leq \exp(-c''' \log n) \leq n^{-c'''},
\end{align*}
where the last inequality follows since if $W_{\max} =np^t_{\max}$, then $\frac{U_{\max}}{W_{\max}} \geq np^e_{\max}$ which, by assumption, is greater than $c_2 \log n$; and, if $W_{\max} =\log n$, then $\frac{U_{\max}}{W_{\max}} \geq \log n$. 
Combining the previous results we obtain
\[
P((d_{T^2E})_i \geq 2 \Delta_{T^2E}) \leq P(\max_{J}\sum_{\theta \in J}T^2E_{\theta}\geq 2\Delta_{T^2E}) + P(\Gamma^C) \leq n^{-c'} +n^{-c''} +n^{-4c'''}+ n^{-c''}.
 \]
Applying the union bound over all indices $i$ we can bound $\max_i (d_{T^2E})_i \leq c_1 \Delta_{T^2E}$ with probability at least $1-n^{-c''}$. Then, from~Equation (\ref{twonorm}) we arrive at the result claimed in the theorem.
\end{proof}

\subsubsection*{Proof of Theorem \ref{ATEE}}

\begin{proof}
For incidental triangles of type $TE^2$, the generating class is one triangle from $G_t$ and two edges from $G_e$. Consequently, we have 
\begin{align*}
E[(d_{TE^2})_{i}] & =E[\sum_{j} \sum_{k} TE^2_{ijk}]\\
& \leq  \sum_{j} \sum_{k} P(\sum_{k_1 \neq k}T_{ijk_1}>0)P(E_{jk}=1)P(E_{ik}=1) \\
& \leq  \sum_{j} \sum_{k} np_{\max}^{t}(p^e_{\max})^2 \leq n^3p^{t}_{\max}(p^e_{\max})^2 \leq  \Delta_{TE^2}.
\end{align*}

Next, let $I_i=\{(TE^2)_{ijk}, j=\{1,\ldots,n\},\,k=\{1,\ldots,n\}\},$ denote the set of all incidentally generated triangles of type $TE^2$ including a vertex $i$. Then, $(TE^2)_{\theta}$ indexed by $\theta=\{i,j,k\}$ is a family of indicator variables with each variable corresponding to an incidentally generated triangle of type $TE^2$. Two different random variables in the family $(TE^2)_{\theta}$ restricted to the set $I_i$ may be dependent in two ways. First, one of the edges $ij$ or $ik$ of the incidental triangle characterized by $TE^2_{ijk}$, may be an edge from $G_e$ and be an edge in the incidental triangle characterized by $(TE^2)_{ijk'}$ for some $k'$ (see Figure~\ref{dependence}(e)). Second, one of the edges $ij$ or $ik$ may have been created by a triangle from $G_t$, with the same triangle being involved in creating the incidental triangle characterized by $(TE^2)_{ij'k'}$ for some $j'$ and $k'$ (see Figure~\ref{dependence}(f)). Note the second possibility also includes the case when the triangles characterized by $(TE^2)_{ijk}$ and $(TE^2)_{ijk'}$ share an edge $ij$ which is created by a triangle from $G_t$. We refer to these two events as $TC_1$ and $TC_2$, respectively. 

With regards to the event $TC_1$, define the following random variable
\[K_{k'}=(1-T_{ijk'})1(\sum_{k''\neq i}T_{jk'k''}>0)1(E_{ik'}=1).
\] 
Conditioned on $TE^2_{ijk}$, each $K_{k'}$ characterizes an incidentally generated triangle in $I_i$ and contributes to the event $TC_1$; for simplicity, 
we let $I_K$ stand for the set of all such variables $K_{k'}$. Then, for any $TE^2_{ijk}$, the number of additional incidentally generated triangles in $I_i$ contributing to the event $TC_1$ is at most $2 \sum_{k'}K_{k'}$ (Figure~\ref{dependence}(e)).

With regards to the event $TC_2$, define the random variable
\[S_{j'k'}=(1-T_{ij'k'})T_{ijj'}E_{ik'}E_{j'k'}.
\] 
Conditioned on $TE^2_{ijk}$, each $S_{j'k'}$ characterizes an incidentally generated triangle in $I_i$ and leads to the event $TC_2$; for simplicity, we let $I_S$ stand for the set of all such variables $S_{j'k'}$s. Then, for any $TE^2_{ijk}$, the number of additional incidentally generated triangles in $I_i$ contributing to the event $TC_2$ is at most $\sum_{j'}\sum_{k'}S_{j'k'}$ (Figure~\ref{dependence}(f)). 

Define a ``good event" as $\Gamma = \Gamma_2 \cap \Gamma_3 \cap \Gamma_4$, where for any $i,j,k$, $\Gamma_2$, $\Gamma_3$ and $\Gamma_4$ may be described as follows:
\begin{align*}
\Gamma_2 & = \{ \text{The number of triangles in } G_t \text{ incident to an edge } ij \text{ is at most } 3W_{\max}=3\max\{np^t_{\max}, \log n \} \},\\
\Gamma_3 & = \{\text{For a side } ij\text{ there are at most } U_{\max}= \max \{n^2p^t_{\max}p^e_{\max},(\log n)^2\} \text{ vertices } k', \\
& \quad \text{ such that the side } ik' \text{ is an edge from  } G_e \text{ and the side } jk' \text{ belongs to a triangle from } G_t\},   \\
\Gamma_4 & = \{\text{Two vertices } \{i,j\} \text{ have at most } 4\tau_{\max}= 4\max \{n(p^e_{\max})^2,(\log n)\} \text{ common neighbors} \{k'\}\}.
\end{align*}

We again apply Proposition \ref{prop1} to $(d_{TE^2})_i$ under the good event $\Gamma$ to obtain an upper bound on $P(\Gamma^C)$.

Under the event $\Gamma_3$, it holds that $2\sum_{k'}K_{k'} \leq 2U_{\max},$ which in turn implies that the number of $T^2E_{\alpha}$ in $I_i$ that depend on $T^2E_{ijk}$ according to the event $TC_1$ is limited to $U_{\max}$. Furthermore, for the event $\Gamma_4$, we have $ \sum_{j'}\sum_{k'}S_{j'k'} \leq 12\tau_{\max}W_{\max}$ which implies that the number of $T^2E_{\alpha}$ in $I_i$ that depend on $T^2E_{ijk}$ according to the event $TC_2$ is limited to $\tau_{\max}W_{\max}$.

Now, define a set $J \subset I_i$ as follows:
\[
J=\{ \theta : \, \max_{\theta_1 \in J} |\theta_2 \in J;\theta_1 \text{ and } \theta_2 \text{ are dependent } | \leq C \},
\]
where $C$ may be found according to
\[C=\sum_{\theta \in J: \theta \sim \beta} TE^2_{\theta} = 2U_{\max} + 12\tau_{\max}W_{\max} \leq 14 \max \{n^2p^t_{\max}p^e_{\max},np^t_{\max}\log n, n(p^e_{\max})^2 \log n, (\log n )^2 \}.
\]

Then, for $t=\mu=\Delta_{TE^2}$,
\begin{align*}
    P(\max \sum_{\theta \in J}TE^2_{\theta}\geq 2\Delta_{TE^2})  
    & \leq \min \bigg \{\exp \left( -\frac{\Delta_{TE^2}^2}{2C(\Delta_{TE^2} + \Delta_{TE^2}/3)}\right), \left(1+\frac{\Delta_{TE^2}}{2\Delta_{TE^2}}\right)^{-2\Delta_{TE^2}/2C} \bigg \}\\
    & = \min \bigg \{\exp \left(- \frac{\Delta_{TE^2}}{ \frac{8}{3}C}\right),2^{-\Delta_{TE^2}/C} \} \\
    & \leq \exp(-c' \log n) \leq n^{-c'},
\end{align*}
where the last inequality follows since if $C =n^2p^t_{\max}p^e_{\max}$, then $\frac{\Delta_{TE^2}}{C} \geq np^e_{\max},$ which is by assumption greater than $c_2 \log n$; if $C=np^t_{\max}\log n$, then $\frac{\Delta_{TE^2}}{C} \geq \frac{(np^e_{\max})^2}{\log n} \geq 
\log n$; and if $C=n(p^e_{\max})^2\log n$, then $\frac{\Delta_{TE^2}}{C} \geq \frac{n^2p^t_{\max}}{\log n} \geq 
\log n$. Finally, if $C=(\log n )^2,$ then $\frac{\Delta_{TE^2}}{C} \geq \log n$.

We bounded the probability $P(\Gamma_3^C)$ in the previous proof, while a bound on $P(\Gamma_4^C)$ is given in Lemma~\ref{TElight}.

Combining the expressions for all previously evaluated bounds, we obtain
\[
P((d_{TE^2})_i \geq 3 \Delta_{TE^2}) \leq P(Z_C \geq 3\Delta_{TE^2}) + P(\Gamma^C) \leq n^{-c'} +n^{-c''} +n^{-4c'''}+ n^{-c''}.
 \]
Taking the union bound over all $i$, we can show that $\max_i (d_{TE^2})_i \leq c_1 \Delta_{TE^2}$ holds with probability at least $1-n^{-c''}$. The claimed result then follows from Equation (\ref{twonorm}).
\end{proof}

\subsubsection*{Proof of Theorem \ref{AT}}
\begin{proof}
We note that under the given assumptions on $p^e_{\max}$ and $p^t_{\max}$ we have the following results:
\[
\sqrt{D_{E^3}} = \max \{n^{3/2} (p^e_{\max})^{5/2},(n^{1/2}(p^e_{\max})^{1/2}(\log n)^{3/2}\} \leq \max \{n^{-\frac{5}{2}\epsilon},\sqrt{\Delta_t}\} = \sqrt{\Delta_{t}},
\]
\begin{align*}
\Delta_{T^3} = \max \{n^{5} (p^t_{\max})^{3},(\log n)^{4}\} &\leq \max \{\sqrt{\Delta_t}n^4(p^t_{\max})^{5/2},\sqrt{\Delta_t}\} \\ &\leq \max \{\sqrt{\Delta_t}n^{-\frac{5}{2}\epsilon},\sqrt{\Delta_t}\} = \sqrt{\Delta_{t}},
\end{align*}
\begin{align*}
\Delta_{T^2E} = \max \{n^4 (p^t_{\max})^2p^e_{\max},(\log n)^4\} &\leq \max \{\sqrt{\Delta_t}n^3(p^t_{\max})^{3/2}p^e_{\max},\sqrt{\Delta_t}\} \\
& \leq \max \{\sqrt{\Delta_t}n^{-\frac{5}{2}\epsilon},\sqrt{\Delta_t}\} = \sqrt{\Delta_{t}},
\end{align*}
\begin{align*}
\Delta_{TE^2} = \max \{n^3 (p^t_{\max})(p^e_{\max})^2,(\log n)^{4}\} &\leq \max \{\sqrt{\Delta_t}n^2(p^t_{\max})^{1/2}(p^e_{\max})^2,\sqrt{\Delta_t}\}, \\
& \leq \max \{\sqrt{\Delta_t}n^{-\frac{5}{2}\epsilon},\sqrt{\Delta_t}\} = \sqrt{\Delta_{t}}.
\end{align*}
Consequently,
\begin{align*}
    \|A_T - E[A_T]\|_2 & \leq \|A_{T^2}-E[A_{T^2}]\|_2 + \|A_{E^3}-E[A_{E^3}]\|_2 +  \|A_{T^3}-E[A_{T^3}]\|_2\\
    & \quad \quad + \|A_{T^2E}-E[A_{T^2E}]\|_2 + \|A_{TE^2}-E[A_{TE^2}]\|_2 \\
    & \leq c(\sqrt{\Delta_{t}} + \sqrt{D_{E^3}} + \Delta_{T^3} + \Delta_{T^2E} + \Delta_{TE^2})\\
    & \leq \tilde{c} \sqrt{\Delta_{t}},
\end{align*}
where $c$ is the maximum of all constants used for bounding the individual matrix terms, and $\tilde{c}$ is another constant that may be easily computed from the previous inequalities.
\end{proof}

\subsubsection*{Proof of Theorem \ref{RT}}

\begin{proof}
First, note that  $E[A_T]=E[A_{T^2}]+E[A_{E^3}]+E[A_{T^2E}]+E[A_{T^3}]+E[A_{TE^2}],$ and all matrices in the sum under the SupSBM model may be written in the form $C((g-h)I_k + y1_k1_k^T)C^T$. Consequently $E[A_T]$ can also be written in the form $C((g-h)I_k + y1_k1_k^T)C^T$. Then, we have $\lambda_{\min}(E[A_T])=\frac{n}{k}(g-h)$ for some $g$ and $h$. Now note that the $(g-h)$ term in $E[A_T]$ is the sum of the corresponding $(g-h)$ terms in the component matrices, all of which are positive due to the community structure of the SupSBM. Hence, the $(g-h)$ term of $E[A_T]$ is going to be greater than the $(g-h)$ term of $E[A_{T^2}]$, so that $\lambda_{\min}(E[A_T]) \geq \lambda_{\min}(E[A_{T^2}])$. This implies that we can replace $\lambda_{\min}(E[A_T])$ with $\lambda_{\min}(E[A_{T^2}])$ in the upper bound of Equation~(\ref{misclus}). We have already computed $\lambda_{\min}(E[A_{T^2}])$ in Equation~(\ref{lambdaPTT}) and the numerator of Equation~(\ref{misclus}) has been upper bounded in Theorem~\ref{AT}. Combining the results, we arrive at the claimed result.
\end{proof}

\subsubsection*{Proof of Theorem \ref{RTT}}

 \begin{proof}
The first inequality is a result of Equation~(\ref{misclus}) which relates the misclustering rate $R_{T^2}$ with $\|A_{T^2}-E[A_{T^2}]\|_2$ and $\lambda_{\min}(E[A_{T^2}])$ through the Davis-Kahan Theorem. The second inequality is obtained by replacing the numerator with the bound from Theorem~\ref{ATT} and the denominator with the result computed in Equation~(\ref{lambdaPTT}).
 \end{proof}

\subsubsection*{Proof of Theorem \ref{tradeoff}}

\begin{proof}
We have the following asymptotic relationship between the two error rates:
\begin{align*}
 \frac{a_t}{n(a_t-b_t)^2} \asymp \frac{a_e/\delta}{\frac{nm^2(a_e-b_e)^2}{\delta^2}} \asymp \frac{\delta}{m^2n}\frac{a_e}{(a_e-b_e)^2}.
\end{align*}
Hence, the error rate obtained by using the information about edges is $\frac{\delta}{m^2n}$ times that of using triangles. Consequently, the error rate is lower for triangle hyperedges if $\frac{\ell}{m^2n} \lesssim 1$ and higher otherwise. 
\end{proof}

\subsubsection*{Proof of Theorem \ref{REEE}}
 \begin{proof}
The first inequality follows from Equation~(\ref{misclus}) that relates the misclustering rate with $\|A_{E^3}-E[A_{E^3}]\|_2$ and $\lambda_{\min}(E[A_{E^3}])$ through the Davis-Kahan Theorem. The second inequality is obtained by replacing the numerator with the bound from Theorem~\ref{ATE} and the denominator with the result in Equation (\ref{lambdaPTE}).
 \end{proof}

\section*{Appendix B}

\subsubsection*{Proof of Lemma \ref{lightpairs}.}
\begin{proof}
Define $u_{ij}=x_iy_j1((i,j)\in L) +x_jy_i1((j,i)\in L)$ for all $i,j=1,\dots,n$. Then,
\[
\sum_{(i,j) \in L} \sum_{k} x_{i}y_{j}(T_{ijk}-E[T_{ijk}])=\sum_{i<j} \sum_{k} (T_{ijk}-p_{ijk})u_{ij}.
\]
Note that each term in the above sum is a zero-mean random variable bounded in absolute value, $|(T_{ijk}-p_{ijk})u_{ij}|\leq 2\sqrt{\Delta_t}/n$. By applying Bernstein's inequality we have
\begin{align*}
    P \left(|\sum_{i<j} \sum_{k \neq (i,j)} (T_{ijk}-p_{ijk})u_{ij}| \geq c_2\sqrt{\Delta_t}\right) & \leq 2 \exp \left(-\frac{\frac{1}{2}c_2^2\Delta_t}{\sum\limits_{i<j} \sum\limits_{k \neq (i,j)} p_{ijk}(1-p_{ijk})u_{ij}^2 + \frac{1}{3}2\frac{\sqrt{\Delta_t}}{n}c\sqrt{\Delta_t}}\right)\\
    & \leq 2 \exp \left(-\frac{\frac{1}{2}c_2^2\Delta_t}{\max_{i,j} (\sum_{k\neq (i,j)} p_{ijk}) \sum u_{ij}^2 + \frac{2}{3}c_2\frac{\Delta_t}{n}}\right) \\
    & \leq 2 \exp \left(-\frac{\frac{1}{2}c_2^2\Delta_t}{ \frac{\Delta_t}{n}(2 + \frac{2c_2}{3})}\right) \\
    & \leq 2 \exp(-\frac{c_2^2}{4+\frac{4c_2}{3}}n),
\end{align*}
where the third inequality follows as a consequence of two observations. First, since $\Delta_t \geq n^2 \max_{i,j,k} p_{ijk}$, we have 
$$\max_{i,j} (\sum_{k\neq (i,j)} p_{ijk}) \leq n  \max_{i,j,k} p_{ijk} \leq \frac{\Delta_t}{n}.$$ 
Second, 
$$\sum_{i,j}u_{ij}^2 \leq 2 \sum_{i,j} (x_i^2y_j^2 ) \leq 2 \|x\|_2^2\|y\|_2^2 \leq 2.$$
From Lemma 5 in~\cite{vershynin2010introduction} regarding the covering number of a sphere, we have $|\mathcal{N}| \leq \exp (n \log 5)$. Hence,  taking the union bound over all possible $x$ and $y$ we obtain
\[
P\left(\sup_{x,y \in \mathcal{N}} |\sum_{(i,j) \in L} \sum_{k} x_{i}y_{j}(T_{ijk}-E[T_{ijk}])| \geq c_2\sqrt{\Delta_t}\right) \leq \exp \left( \left (-\frac{c_2^2}{4+\frac{4c_2}{3}}+ \log 5 \right)n \right).
\]
The claimed result now follows from selecting a sufficiently large constant $c_2$ and $r_1=(-\frac{c_2^2}{4+\frac{4c_2}{3}}+ \log 5 )$.
\end{proof}

\subsubsection*{Proof of Lemma \ref{heavypairs}.}
\begin{proof}
We first address the subset of heavy pairs $H_1= \{(i,j) \in H: x_i >0, y_j>0\}$. The other cases may be analyzed similarly. 

Define the following two families of sets:
\[
I_1=\{\frac{2^{-1}}{\sqrt{n}} \leq x_i \leq \frac{1}{\sqrt{n}} \}, \quad  I_s=\{\frac{2^{s-1}}{2\sqrt{n}} < x_i \leq \frac{2^{s}}{2\sqrt{n}}\}, \,s=2,3,\ldots, \lceil \log_2 2\sqrt{n} \rceil,
\]
\[
J_1=\{\frac{2^{-1}}{\sqrt{n}} \leq y_i \leq \frac{1}{\sqrt{n}} \}, \quad  J_t=\{\frac{2^{t-1}}{2\sqrt{n}} < y_i \leq \frac{2^{t}}{2\sqrt{n}}\}, \, t=2,3,\ldots, \lceil \log_2 2\sqrt{n} \rceil.
\]
Next, for two arbitrary sets $I$ and $J$ of vertices, also define
\[
e(I,J)=\begin{cases}
\sum_{i \in I} \sum_{j \in J} \sum_{k \neq (i,j)} T_{ijk} & I \cap J =\emptyset, \\
\sum_{(i,j) \in I \times J \backslash (I \cap J)^2} \sum_{k \neq (i,j)} T_{ijk} + \sum_{(i,j) \in  (I \cap J)^2, i<j} \sum_{k \neq (i,j)} T_{ijk}  & I \cap J \neq \emptyset,
\end{cases}
\]
\[
\mu(I,J)=E[e(I,J)], \quad \bar{\mu}=|I||J| n\max_{i,j,k} p_{ijk} \leq |I||J| \frac{\Delta_t}{n},
\]
Finally, let $\bar{\mu}_{st}=\bar{\mu}(I_s,J_t)$ , $\lambda_{st} =e(I_s,J_t)/\bar{\mu}_{st}$, $\alpha_s =|I_s|2^{2s}/n$, $\beta_t=|J_t|2^{2t}/n$, and $\sigma_{st}=\lambda_{st}\sqrt{\Delta_t}2^{-(s+t)}$.

We have the following two results establishing relationships between the previously introduced entities.

\begin{lem}
Let $d_{t,i} = \sum_{j} \sum_{k \neq i,j} T_{ijk}$ denote the triangle-degree of vertex $i$. Then, for all $i$, and a constant $r_3>0$, there exists a constant $c_4(r_3)>0 $ such that $d_{t,i} \leq c_4\Delta_t$ with probability at least $1-n^{-r_3}$.
\label{bounded degree}
\end{lem}

\begin{lem}
For a constant $r_4>0$, there exists constants $c_5(r_4), c_6(r_4)>1$ such that for any pair of vertex sets $I,J \subseteq \{1,\ldots,n\}$ such that $|I|\leq |J|$, with probability at least $1-2n^{-r_4}$, at least one of the following statements holds:
\vspace{5pt}

(a) $
\frac{e(I,J)}{\bar{\mu} (I,J)} \leq e\, c_5,$

(b) $
e(I,J) \log \frac{e(I,J)}{\bar{\mu} (I,J)} \leq c_6\, |J| \, \log \frac{n}{|J|}.$
\label{twostatement}
\end{lem}

Now, we use the result of the two previous lemmas to complete the proof of the claimed result for the heavy pairs. We note
\begin{align*}
    \sum_{(i,j) \in H_1} x_i y_j \sum_{k \neq (i,j)} T_{ijk}  \leq 2 \sum_{(s,t): 2^{(s+t)}\geq \sqrt{\Delta_t}} e(I_s,J_t)\frac{2^s}{2\sqrt{n}}\frac{2^t}{2\sqrt{n}}  \leq \frac{\sqrt{\Delta_t}}{2} \sum_{(s,t): 2^{(s+t)}\geq \sqrt{\Delta_t}} \alpha_s \beta_t \sigma_{st}.
\end{align*}

We would like to bound the right-hand-side of the inequality by a constant multiple of $\sqrt{\Delta_t}$. To this end, first note the following two facts:
\[
\sum_{s}\alpha_s \leq 4 (1/2)^{-2} =1, \quad \sum_{t}\beta_{t} \leq 1.
\]
Following the approach of ~\cite{lei2015consistency} and~\cite{ chin2015stochastic}, we split the set of pairs $C: \{(s,t): 2^{(s+t)}\geq \sqrt{\Delta_t}, |I_s| \leq |J_t| \}$ into six parts and show that desired invariant for each part is bounded. 
\begin{itemize}
   \item  $C_1: \{(s,t) \in C, \sigma_{st} \leq 1\}$: 
    \[
    \sum_{(s,t)}\alpha_s \beta_t \sigma_{st}1\{(s,t) \in C_1 \} \leq \sum_{s,t}\alpha_s \beta_t \leq 1.
    \]
    
    \item $C_2: \{(s,t) \in C\backslash C_1, \lambda_{st} \leq e\, c_5\}$: \\
    Since
    \[
    \sigma_{st}=\lambda_{st}\sqrt{\Delta_t}2^{-(s+t)} \leq \lambda_{st} \leq e\, c_5, 
    \]
    consequently
    \[
    \sum_{(s,t)}\alpha_s \beta_t \sigma_{st}1\{(s,t) \in C_2 \} \leq e \, c_5 \sum_{s,t}\alpha_s \beta_t \leq e\, c_5.
    \]
    \item $C_3: \{(s,t) \in C\backslash (C_1 \cup C_2), 2^{s-t} \geq \sqrt{\Delta_t}\}$:\\
    By Lemma (\ref{bounded degree}), $e(I_s,J_t)\leq c_4 |I_s|\Delta_t$. Hence, 
    $$\lambda_{st} =e(I_s,J_t)/\bar{\mu}_{st} \leq c_4\frac{|I_s|\Delta_t}{|I_s||J_t|\Delta_t/n} \leq c_4\frac{n}{|J_t|},$$ 
    and consequently, 
    $$\sigma_{st} \leq c_4\sqrt{\Delta_t}2^{-(s+t)}\frac{n}{|J_t|} \leq c_42^{-2t}\frac{n}{|J_t|},$$ 
    for $(s,t) \in C_3$.
    Then,
    \begin{align*}
    \sum_{(s,t)}\alpha_s \beta_t \sigma_{st}1\{(s,t) \in C_3 \} & \leq \sum_s \alpha_s \sum_t \beta_t c_1\, 2^{-2t}\frac{n}{|J_t|} \\
    & \leq  \sum_s \alpha_s \sum_t 2^{2t}\frac{|J_t|}{n} c_4\, 2^{-2t}\frac{n}{|J_t|} \leq c_4\, \sum_s \alpha_s \leq c_4.
    \end{align*}
    \item $C_4: \{(s,t) \in C\backslash (C_1 \cup C_2 \cup C_3), \log \lambda_{st} > \frac{1}{4}[2t \log 2 + \log (1/\beta_t)] \}$:\\
    From part (b) of Lemma \ref{twostatement}, we have,
    \[
    \lambda_{st} \log \lambda_{st} \frac{|I_s| |J_t| \Delta_t}{n} \leq \frac{e(I_s,J_t)}{\bar{\mu}(I_s,J_t)} \log \frac{e(I_s,J_t)}{\bar{\mu}(I_s,J_t)} \bar{\mu}(I_s,J_t) \leq  c_6\, |J_t| \log \frac{2^{2t}}{|J_t|},
    \]
    which is equivalent to
    \[
        \sigma_{st} \alpha_s  \leq  c_6 \frac{1}{\log \lambda_{st} }\frac{2^{s-t}}{\sqrt{\Delta_t}}\{2t \log 2 + \log (1/\beta_t)\} \leq 4\,c_6 \frac{2^{s-t}}{\sqrt{\Delta_t}}.
    \]
    Then,
    \begin{align*}
    \sum_{(s,t)}\alpha_s \beta_t \sigma_{st}1\{(s,t) \in C_4 \} & = \sum_t \beta_t \sum_s  \sigma_{st} \alpha_s 1\{(s,t) \in C_4 \} \\
    & \leq 4\, c_6 \sum_t \beta_t \sum_s \frac{2^{s-t}}{\sqrt{\Delta_t}} 1\{(s,t) \in C_4 \} \leq 8\, c_6 \sum_t \beta_t \leq 8\, c_6.
    \end{align*}
    \item $C_5: \{(s,t) \in C\backslash (C_1 \cup C_2 \cup C_3 \cup C_4), 2t \log 2 \geq \log (1/\beta_t)] \}$:\\
    First, note that since $(s,t) \notin C_4$, we have $\log \lambda_{st} \leq \frac{1}{4}[2t \log 2 + \log (1/\beta_t)] \leq t \log 2$ and hence $\lambda_{st} \leq 2^t$. Next, $\sigma_{st} =\lambda_{st} \sqrt{\Delta_t}2^{-(s+t)} \leq 2^{-s}\sqrt{\Delta_t},$ and hence
    $\sigma_{st}\alpha_s \leq 4c_6 \frac{2^{s-t}}{\sqrt{\Delta_t}} 4t \log 2$.   
    Therefore,
    \[
    \sum_{(s,t)}\alpha_s \beta_t \sigma_{st}1\{(s,t) \in C_5 \} \leq \sum_t \beta_t \sum_s 4\, c_6 \frac{2^{s-t}}{\sqrt{\Delta_t}} 4t \log 2 \leq 2\, c_6 \log 2 \sum_t \beta_t \leq 2\, c_6.
    \]     
    \item $C_6: \{(s,t) \in C\backslash (C_1 \cup C_2 \cup C_3 \cup C_4 \cup C_5) \}$:\\
    Since $2t\log 2 < \log (1/\beta_t)$, we have $\log \lambda_{st} \leq t \log 2 \leq \log (1/\beta_t) /2$. This observation, along with the fact $\lambda_{st} \geq 1$, implies that $\lambda_{st} \leq 1/\beta_t$.
     As a result,
    \[
    \sum_{(s,t)}\alpha_s \beta_t \sigma_{st}1\{(s,t) \in C_6 \} \leq  \sum_s \alpha_s \sum_t  2^{-(s+t)}\sqrt{\Delta_t} \{(s,t) \in C_6 \} \leq \sum_s \alpha_s \leq 2.
    \]
\end{itemize}
In a similar fashion, the set of pairs $C: \{(s,t): 2^{(s+t)}\geq \sqrt{\Delta_t}, |I_s| > |J_t| \}$ is split into six categories in order to bound $\sum_{(s,t)} \alpha_s \beta_t \sigma_{st}$. The derivations are omitted. 

Collecting all the previously obtained terms, we arrive at the claimed result for heavy pairs: for some constant $r_2>0$, there exists a constant $c_3(r_2)>0$ such that with probability at least $1-2n^{-r_2}$, one has
\[
\sum_{(i,j) \in H} \sum_{k} x_{i}y_{j}T_{ijk} \leq c_3 \sqrt{\Delta_t}.
\]
\end{proof}

\subsubsection*{Proof of Lemma \ref{TElight}}
\begin{proof}
As before, define $u_{ij}=x_iy_j1((i,j)\in L) +x_jy_i1((j,i)\in L)$ for all $i,j=1,\dots,n$. Then,
\[
\sum_{(i,j) \in L} x_{i}y_{j}(A_{E^3}- E[A_{E^3}])_{ij}=\sum_{i<j} \sum_{k \neq (i,j)} (E_{ij}E_{jk}E_{ik} - p^e_{ij}p^e_{jk}p^e_{ik})u_{ij}.
\]
To analyze the above sum, we use the \emph{typical bounded differences inequality} of Theorem 1.3 in~\cite{warnke2016method}. 
For this purpose, define 
$$f(E) = \sum_{i<j} \sum_{k \neq (i,j)} E_{ij}E_{jk}E_{ik} u_{ij}.$$ 
Clearly, $f$ is a low-order polynomial of independent random variables. More precisely, since $E_{ij}$ are independent Bernoulli  random variables with parameters $p^e_{ij}$, we have 
$$E[f(A)]=\sum_{i<j} \sum_{k \neq (i,j)}p^e_{ij}p^e_{jk}p^e_{ik}u_{ij}.$$ 
Let $\tau_{ij}$ be the number of common neighbors of the vertices $i$ and $j$, i.e., $\tau_{ij}=\sum_{k \neq {i,j}}E_{ik}E_{jk}$. Then $\tau_{ij}$ is a sum of $n-2$ independent Bernoulli random variables with parameters $p^e_{ik}p^e_{jk} \leq (p^e_{\max})^2$.
Next, define a ``good set" $\Gamma$ under which the contribution of one single random variable to the function is limited with high probability as follows:
\[
\Gamma=\{(E_{ij}): \max_{ij} \tau_{ij} \leq 4\tau_{\max}\},
\]
asserting that every pair of vertices $i,j$ has at most $4\tau_{\max}$ common neighbors.
From Bernstein inequality we have,
\begin{align*}
    P(E \notin \Gamma) & \leq n^2 P (\tau_{ij} > 4\tau_{max}) \\
    & \leq n^2 P \left(\sum_{k \neq {i,j}}(E_{ik}E_{jk} -p^e_{ik}p^e_{jk}) > 3\tau_{max}\right) \\
    & \leq n^2 \exp \left(-\frac{9\tau^2_{\max} }{2\sum_{k}p^e_{ik}p^e_{jk}(1-p^e_{ik}p^e_{jk})+\frac{3}{3}\tau_{\max}}\right) \\
    & \leq n^2 \exp(-\frac{9\tau^2_{\max}}{2\tau_{\max}+\tau_{\max}}) \\
    &\leq n^2 \exp(-\frac{9}{3}\tau_{\max}) \leq \exp(-\log n),
\end{align*}
where the last inequality holds since $\tau_{\max}>\log n$ by definition. Observe that the good event $\Gamma$ is independent on the particular values of $x,y$.  

Next, we determine the typical Lipschitz (TL) condition for the function $f(E)$. Changing one element in the sequence $(E_{11},E_{12},\ldots,E_{nn})$ (say, $E_{ij}$) from $1$ to $0$ may have two different types of effects on $f(E)$. The effect may be ``large" on the term $(A_{E^3})_{ij}u_{ij}= \sum_{k \neq (i,j)} E_{ij}E_{jk}E_{ik} u_{ij}$, which is upper bounded by $\sum_{k \neq (i,j)} E_{jk}E_{ik} u_{ij}$. 
Or, the effects may be ``small" for the terms of the form $(A_{E^3})_{ik}u_{ik}$ and $(A_{E^3})_{jk}u_{jk}$ (i.e., the terms that involve the entries of the matrix $A_{E^3}$ which represent neighboring edges of $(i,j)$). These ``small'' effects are upper bounded by $u_{ik}$ and $u_{jk}$, respectively.

Under the good event $\Gamma$, we have $\sum_{k \neq i,j} E_{ik}E_{jk} \leq 4\tau_{\max}$, and consequently, the ``large'' effect of changing $E_{ij}$ from 1 to 0 is bounded by $4\tau_{\max}u_{ij}$. Moreover, under the good event $\Gamma$, there are at most $4\tau_{\max}$ ``small'' effects, since for a ``small'' effect to occur, both $E_{ik}$ and $E_{jk}$ must be 1, i.e., $ik$ and $jk$ must be connected by an edge. However, an additional complication is that the effects contribute differently to the bound, depending upon which common neighbor $k$ we are looking at. To mitigate this problem, we first lump the ``small'' effects together into $\sum_{k:\, E_{ik}E_{jk}=1}u_{ik}$. Combining the ``large'' and ``small'' effects, under the good event,  
$$c_{ij}=4\tau_{\max}u_{ij}+\sum_{k: E_{ik}E_{jk}=1, E\in \Gamma}u_{ik}$$ 
emerges as an upper bound on the total effect of changing one $E_{ij}$ in $f(E)$. For the case that the bad event occurs instead, an upper bound on the effect of the change is $d_{ij}=2nu_{ij}+\sum_{k: E_{ik}E_{jk}=1}u_{ik}$. 

Now, let $\gamma_{ij}=\frac{1}{n},$ for all $i,j$. Then $e_{ij}=o(c_{ij})$, and
\[
C=\max_{ij}c_{ij}=4\tau_{\max}\frac{\sqrt{D_{E^3}}}{n\tau_{\max}} +  4\tau_{\max} \frac{\sqrt{D_{E^3}}}{n\tau_{\max}} = 8\tau_{\max}\frac{\sqrt{D_{E^3}}}{n\tau_{\max}} =8\frac{\sqrt{D_{E^3}}}{n}.
\]
Next, we need to compute an upper bound on $\sum_{ij}c_{ij}^2$. This can be done as follows
\begin{align*}
c_{ij}^2 & \leq  2(16\tau^2_{\max}u^2_{ij} + (\sum_{k: a_{ik}a_{jk}=1, a \in \Gamma}u_{ik})^2),  \quad \quad \quad \quad \text{(since $(a+b)^2 \leq 2(a^2+b^2)$)} \\
& \leq 2(16\tau^2_{\max}u^2_{ij} + (\sum_{k: a_{ik}a_{jk}=1, a \in \Gamma} 1) (\sum_{k: a_{ik}a_{jk}=1, a \in \Gamma} u^2_{ik})) \quad \quad \text{(due to the Cauchy-Schwartz inequality)}\\
& \leq 2(16\tau^2_{\max}u^2_{ij} + 4\tau_{\max} (\sum_{k: a_{ik}a_{jk}=1, a \in \Gamma} u^2_{ik})).
\end{align*}
Within the sum of $c_{ij}^2$ over all $i,j$, each term $u_{ik}^2$ will appear at most $\sum_{k: a_{ik}a_{jk}=1,a \in \Gamma}=4\tau_{\max}$ times. This implies that $\sum_{ij}c^2_{ij} \leq 64 \tau^2_{\max} \sum_{ij}u_{ij}^2$.

In the notations of  Theorem 1.3 of~\cite{warnke2016method} define $\gamma_{ij} =\frac{1}{n}$ for all $ij$ and the event $\mathcal{B}(\Gamma,\{\gamma_{n}\})$, such that $\mathcal{B}^C \subset \Gamma$ and
\[
P(\mathcal{B}^{C}) \leq P(A \notin \Gamma) \sum_{i,j}\gamma^{-1}= n^3 P(A \notin \Gamma) = \exp(-(c'-3)\log n).
\]
Then using Theorem 1.3 of~\cite{warnke2016method}, we have
\begin{align*}
    P(\{{f(A)-E[f(A)] \geq \sqrt{D_{TE}}\}} \, \cap \,\mathcal{B}^{C}) & \leq \exp(- \frac{D_{E^3}}{2 \sum_{ij}p_{ij}(1-p_{ij})(c_{ij} + e_{ij})^2 + 2C\sqrt{D_{E^3}}/3})  \\
    & \leq \exp(- \frac{D_{E^3}}{2 \sum_{ij}p_{\max}(64\tau^2_{\max})u^2_{ij} + \frac{16}{3}\frac{\sqrt{D_{E^3}}}{n}\sqrt{D_{E^3}}})  \\
    & \leq \exp(- \frac{D_{E^3}}{128p_{\max}\tau_{\max}^2 + \frac{16}{3}\frac{\sqrt{D_{E^3}}}{n}\sqrt{D_{E^3}}}) \\
    & \leq \exp(- \frac{D_{E^3}}{(128 + 16/3)\frac{D_{E^3}}{n}}) \leq \exp(-cn),
\end{align*}
where the penultimate inequality follows since $D_{E^3}=np^e_{\max}\tau_{\max}^2$.

Clearly, the event $\{A \notin \Gamma\}$ does not depend on the choice of the vectors $x,y$. Hence, taking the supremum over all $x$ and $y$, we have
\[
P(\sup_{x,y \in \mathcal{N}} |\sum_{i,j \in L} x_iy_j(A_{E^3}-E[A_{E^3}])_{ij}\geq \sqrt{D_{E^3}}) \leq \exp(-(c-\log 5)n) + \exp(-(c'-3)\log n).
\]
This completes the proof.
\end{proof}

\subsubsection*{Proof of Lemma \ref{TEheavy1}}
\begin{proof}
As before, we first focus on the subset of heavy pairs, $H_1=\{(i,j) \in H: x_i >0, y_j>0\}$; the other two cases follow similarly. The vertex sets $I_1,\ldots,I_{\lceil\log_2 2\sqrt{n}\rceil}, J_1, \ldots, J_{\lceil\log_2 2\sqrt{n} \rceil}$ are defined as before. In addition, we write
\[
e(I,J)=\begin{cases}
\sum_{i \in I} \sum_{j \in J}(A_{E^3})_{ij} & I \cap J =\emptyset \\
\sum_{(i,j) \in I \times J \backslash (I \cap J)^2} (A_{E^3})_{ij} + \sum_{(i,j) \in  (I \cap J)^2, i<j} (A_{E^3})_{ij}  & I \cap J \neq \emptyset
\end{cases},
\]
\[
\mu(I,J)=E(e(I,J)), \quad \bar{\mu}=|I||J| n(p^e_{\max})^3,
\]
\[ \bar{\mu}_{st}=\bar{\mu}(I_s,J_t) , \lambda_{st} =e(I_s,J_t)/\bar{\mu}_{st}, \alpha_s =|I_s|2^{2s}/n, \beta_t=|J_t|2^{2t}/n, \text{ and } \sigma_{st}=\lambda_{st}\frac{\sqrt{D_{TE}}}{\tau_{\max}}2^{-(s+t)}.\]

 The degree of row $i$ of the matrix $A_{E^3}$ is $(d_{E^3})_i=\sum_{j} (A_{E^3})_{ij}=\sum_{j} \sum_{k}E_{ij}E_{jk}E_{ik}$. Hence, $(d_{E^3})_i$ counts the number of triangles incident to the vertex $i$. Let $\Delta_{TE} = np_{\max}T_{\max} = \max \{ n^2 p_{\max}^3, np_{\max} \log n\}$. Then $\Delta_{TE}$ may be vaguely interpreted as being the (approximate) maximum degree of the rows of the $P_{TE}$ matrix. The next lemma bounds the degrees of the rows of the matrix $A_{TE}$ with high probability. 
\begin{lem}
If $np^e_{\max} > \log n$, then for a constant $r_3>0$, there exists a constant $c_4(r_3)>0 $ such that the ``degree" of row $i$, $(d_{E^3})_i \leq c_4\Delta_{E^3}$ with probability at least $1-n^{-r_3}$ for all $i$.
\label{bounded degree TE}
\end{lem}

\begin{lem}
For a constant $c>0$, there exist constants $c_2(c), c_3(c)>1$ such that with probability at least $1-2n^{-c}$ and for any vertex sets $I,J \subseteq [n]$ and $|I|\leq |J|$ one of the following two statements is true:

\vspace{5pt}
(a) $\frac{e(I,J)}{\bar{\mu} (I,J)} \leq ec_2,$

(b) $
e(I,J) \log \frac{e(I,J)}{\bar{\mu} (I,J)} \leq c_3 n(p^e_{\max})^2 |J| \log \frac{n}{|J|}.$
\label{two statement TE}
\end{lem}

We use the result of the two previous lemmas to establish the proof for heavy pairs. In this setting, note that
\begin{align*}
    \sum_{(i,j) \in H_1} x_i y_j (A_{E^3})_{ij}  & \leq 2 \sum_{(s,t): 2^{(s+t)}\geq \frac{\sqrt{D_{E^3}}}{\tau_{\max}}} e(I_s,J_t)\frac{2^s}{2\sqrt{n}}\frac{2^t}{2\sqrt{n}} \\
    &= 2 \, \frac{1}{4} \sum_{(s,t): 2^{(s+t)}\geq \frac{\sqrt{D_{E^3}}}{\tau_{\max}}} \frac{e(I_s,J_t)}{|I_s||J_t|(np^e_{\max})^3}\frac{D_{E^3}}{\tau_{\max}}\frac{2^s|I_s|2^t|J|_t}{n^2} \\
    & =  \frac{1}{2} \sqrt{D_{E^3}}\sum_{(s,t): 2^{(s+t)}\geq \frac{\sqrt{D_{E^3}}}{\tau_{\max}}} \frac{e(I_s,J_t)}{\bar{\mu}(I_s,J_t)}\frac{\sqrt{D_{E^3}}}{\tau_{\max}} 2^{-(s+t) }\frac{2^{2s}|I_s|2^{2t}|J|_t}{n^2} \\
    & = \frac{\sqrt{D_{E^3}}}{2} \sum_{(s,t): 2^{(s+t)}\geq \frac{\sqrt{D_{E^3}}}{\tau_{\max}}} \alpha_s \beta_t \sigma_{st}.
\end{align*}
Next, we need to bound this quantity by a constant multiple of $\sqrt{D_{E^3}}$. Following the approach of~\cite{lei2015consistency} and~\cite{ chin2015stochastic}, we split the set of pairs $C: \{(s,t): 2^{(s+t)}\geq \frac{\sqrt{D_{E^3}}}{\tau_{\max}}, |I_s| \leq |J_t| \}$ into six parts and show that the contribution of each part is bounded accordingly. Again, in our proof we rely on two facts,
\[
\sum_{s}\alpha_s \leq \sum_{i}|4x_i|^2 \leq 16, \quad \sum_{t}\beta_{t} \leq 16.
\]
\begin{itemize}
   \item  $C_1: \{(s,t) \in C, \sigma_{st} \leq 1\}$: 
    \[
    \sum_{(s,t)}\alpha_s \beta_t \sigma_{st}1\{(s,t) \in C_1 \} \leq \sum_{s,t}\alpha_s \beta_t \leq 256.
    \]
    
    \item $C_2: \{(s,t) \in C\backslash C_1, \lambda_{st} \leq c_3\,e\}$:\\
    Under $C$, $2^{s+t} \geq \frac{\sqrt{D_{E^3}}}{\tau_{\max}}$, and consequently,
    \[
    \sigma_{st}=\lambda_{st}\frac{\sqrt{D_{E^3}}}{\tau_{\max}}2^{-(s+t)} \leq \lambda_{st} \leq c_3\, e. 
    \]
    This implies
    \[
    \sum_{(s,t)}\alpha_s \beta_t \sigma_{st}1\{(s,t) \in C_2 \} \leq c_3\, e\, \sum_{s,t}\alpha_s \beta_t \leq 256\,c_3\,e.
    \]
    \item $C_3: \{(s,t) \in C\backslash (C_1 \cup C_2), 2^{s-t} \geq \frac{\sqrt{D_{E^3}}}{\tau_{\max}}\}$:\\
    By Lemma (\ref{bounded degree TE}), $e(I_s,J_t)\leq c_1 |I_s|\Delta_{E^3}$. Hence, 
    $$\lambda_{st} =e(I_s,J_t)/\bar{\mu}_{st} \leq c_1\frac{|I_s|\Delta_{E^3}}{|I_s||J_t|(np^e_{\max})^3} \leq c_1\frac{n}{|J_t|},$$ 
    and consequently, 
    \[\sigma_{st} = \lambda_{st}\frac{\sqrt{D_{E^3}}}{\tau_{\max}}2^{-(s+t)}\leq c_1 \frac{n}{|J_t|}\frac{\sqrt{D_{E^3}}}{\tau_{\max}}2^{-(s+t)} \leq c_12^{-2t}\frac{n}{|J_t|}, 
    \]
    for $(s,t) \in C_3$. Then,
    \begin{align*}
    \sum_{(s,t)}\alpha_s \beta_t \sigma_{st}1\{(s,t) \in C_3 \} & \leq \sum_s \alpha_s \sum_t \beta_t c_12^{-2t}\frac{n}{|J_t|} \\
    & \leq  \sum_s \alpha_s \sum_t 2^{2t}\frac{|J_t|}{n} c_12^{-2t}\frac{n}{|J_t|} \leq c_1\sum_s \alpha_s \leq 256\, c_1.
    \end{align*}
    
    \item $C_4: \{(s,t) \in C\backslash (C_1 \cup C_2 \cup C_3), \log \lambda_{st} > \frac{1}{4}[2t \log 2 + \log (1/\beta_t)] \}$:\\
    From part (b) of Lemma~\ref{twostatement}, we have
    \[
    \lambda_{st} \log \lambda_{st} |I_s| |J_t| np^3_{\max} \leq \frac{e(I_s,J_t)}{\bar{\mu}(I_s,J_t)} \log \frac{e(I_s,J_t)}{\bar{\mu}(I_s,J_t)} \bar{\mu}(I_s,J_t) \leq  c_4 T_{\max}|J_t| \log \frac{n}{|J_t|}.
    \]
Noting that $\frac{\tau_{\max}}{n^2(p^e_{\max})^3}= \frac{\tau_{\max}^2}{n^2(p^e_{\max})^3\tau_{\max}}=\frac{\tau_{\max}^2}{D_{E^3}}$, we may write
    \begin{align*}
        \sigma_{st} \alpha_s  & \leq \lambda_{st}\frac{\sqrt{D_{E^3}}}{\tau_{\max}}2^{-(s+t)} \frac{|I_s|2^{2s}}{n} \\
        & \leq c_4 \frac{1}{\log \lambda_{st} } \frac{(\tau_{\max})^2}{D_{E^3}}\log (\frac{n}{|J_t|}) \frac{\sqrt{D_{E^3}}}{\tau_{\max}}2^{(s-t)}
        \\
        &\leq  c_4 \frac{1}{\log \lambda_{st} }\frac{2^{s-t}}{\frac{\sqrt{D_{E^3}}}{\tau_{\max}}}\{2t \log 2 + \log (1/\beta_t)\} \leq 4\, c_4 \, \frac{2^{s-t}}{\sqrt{D_{E^3}}/\tau_{\max}}.
    \end{align*}
    Then,
    \begin{align*}
    \sum_{(s,t)}\alpha_s \beta_t \sigma_{st}1\{(s,t) \in C_4 \} & = \sum_t \beta_t \sum_s  \sigma_{st} \alpha_s 1\{(s,t) \in C_4 \} \\
    & \leq 4\, c_4\, \sum_t \beta_t \sum_s \frac{2^{s-t}}{\sqrt{D_{E^3}}/\tau_{\max}} 1\{(s,t) \in C_4 \} \leq 8\, c_4 \sum_t \beta_t \leq 128\, c_4,
    \end{align*}
    where the penultimate inequality relies on the fact that $(s,t) \notin C_3$, and that consequently $2^{(s-t)} \leq \frac{\sqrt{D_{E^3}}}{\tau_{\max}}$.
    \item $C_5: \{(s,t) \in C\backslash (C_1 \cup C_2 \cup C_3 \cup C_4), 2t \log 2 \geq \log (1/\beta_t) \}$:\\   
    First, note that since $(s,t) \notin C_4$, we have $\log \lambda_{st} \leq \frac{1}{4}[2t \log 2 + \log (1/\beta_t)] \leq t \log 2$ and hence $\lambda_{st} \leq 2^t$. Furthermore, $\sigma_{st} =\lambda_{st}\frac{\sqrt{D_{E^3}}}{\tau_{\max}}2^{-(s+t)} \leq 2^{-s}\frac{\sqrt{D_{E^3}}}{\tau_{\max}}$. Since $(s,t) \notin C_2$, we have $\log \lambda_{st} \geq 1$ and
    \[\sigma_{st}\alpha_s \leq c_4 \frac{1}{\log \lambda_{st} }\frac{2^{s-t}}{\frac{\sqrt{D_{E^3}}}{\tau_{\max}}}\{2t \log 2 + \log (1/\beta_t)\} \leq c_4 \frac{2^{s-t}}{\frac{\sqrt{D_{TE}}}{\tau_{\max}}} 4\,t \, \log 2.
    \]    
    Then,
    \[
    \sum_{(s,t)}\alpha_s \beta_t \sigma_{st}1\{(s,t) \in C_5 \} \leq \sum_t \beta_t \sum_s c_4 \frac{2^{s-t}}{\frac{\sqrt{D_{E^3}}}{\tau_{\max}}} 4t \log 2 \leq 2\, c_4 \, \log 2 \sum_t \beta_t \leq 32\, c_4.
    \]
    \item $C_6: \{(s,t) \in C\backslash (C_1 \cup C_2 \cup C_3 \cup C_4 \cup C_5) \}$:\\
    Since $2t\log 2 < \log (1/\beta_t)$, we have $\log \lambda_{st} \leq t \log 2 \leq \log (1/\beta_t) /2$. This fact, along with $\lambda_{st} \geq 1,$ implies that $\lambda_{st} \leq 1/\beta_t$. Therefore,
    \[
    \sum_{(s,t)}\alpha_s \beta_t \sigma_{st}1\{(s,t) \in C_6 \} \leq  \sum_s \alpha_s \sum_t  2^{-(s+t)}\frac{\sqrt{D_{E^3}}}{\tau_{\max}} \{(s,t) \in C_6 \} \leq \sum_s \alpha_s \leq 16.
    \]
\end{itemize}
The set of pairs $C: \{(s,t): 2^{(s+t)}\geq \sqrt{\Delta_t}, |I_s| > |J_t| \}$ may be similarly split into six categories categories and similar arguments may be used to bound each of the contributions $\sum_{(s,t)} \alpha_s \beta_t \sigma_{st}$. Collecting all the terms we have the following result for heavy pairs: for some constant $c>0$, there exists a constant $c_1(c)>0$ such that with probability at least $1-2n^{-c}$, one has 
\[
\sum_{(i,j) \in H}  x_{i}y_{j}(A_{E^3})_{ij} \leq c_1\sqrt{D_{E^3}}.
\]
\end{proof}

\subsubsection*{Proof of Lemma \ref{bounded degree}.}
\begin{proof}
We note $d_{t,i}=\sum_{j}\sum_{k} T_{ijk}$ is a sum of independent random variables, each bounded in absolute value by 1. Therefore, Bernstein's inequality gives
\begin{align*}
    P(d_{t,i} \geq c_4\Delta_t) & \leq P\left(\sum_{j}\sum_{k} w_{ijk} \geq (c_4-1)\Delta_t\right) \\
    & \leq \exp \left( -\frac{\frac{1}{2}(c_4-1)^2\Delta_t^2}{\sum_j \sum_k p_{ijk}(1-p_{ijk}) + \frac{1}{3}(c_4-1)\Delta_t}\right)\\
    & \leq \exp (-\Delta_t \frac{3(c_4-1)^2}{2c_4 + 4}) \\
    & \leq n^{-c_7},
\end{align*}
where the last inequality follows since $\Delta_t \geq c \log n$. Taking the union bound over all values of $i$ we obtain that $\max_i d_{t,i} \leq c_4 \Delta_t$ with probability at least $1-n^{-r_3},$ where $c_4$ is a function of the constant $r_3$. 
\end{proof}

\subsubsection*{Proof of Lemma \ref{twostatement}.}
\begin{proof}
If $|J|>n/e$, then the result of Lemma \ref{bounded degree} implies 
\[
    \frac{e(I,J)}{\Delta_t |I||J|/n}  \leq \frac{\sum_{i \in I} \max_{i} d_{t,i}}{\Delta_t |I|/e} \leq \frac{|I|c_2\Delta_t}{\Delta_t|I|/e} \leq c_2\, e,
\]
and consequently, (a) holds for this case.

If $|J| < n/e$, let $S(I,J)=\{(i,j),i \in I, j \in J\}$. We next invoke Corollary A.1.10 of~\cite{alon2004probabilistic}, described below.
\begin{prop}
For independent Bernoulli random variables $X_u \sim Bern(p_u), u=1,\ldots,n$ and $p=\frac{1}{n}\sum_{u}p_{u}$, we have
\[
P(\sum_{u} (X_u -p_u) \geq a) \leq \exp (a- (a+ pn) \log (1+ a/pn)).
\]
\end{prop}
Using the above result, for $l\geq 8$, we have
\begin{align*}
    P(e(I,J) \geq l \bar{\mu}(I,J)) & \leq P (\sum_{(i,j) \in S(I,J)}\sum_{k\neq (i,j)} (T_{ijk}-p_{ijk}) \geq l \bar{\mu}(I,J) - \sum_{(i,j) \in S(I,J)} \sum_{k\neq (i,j)} p_{ijk} ) \\
    & \leq P (\sum_{(i,j) \in S(I,J)}\sum_{k\neq (i,j)} w_{ijk} \geq (l-1) \bar{\mu}(I,J) ) \\
    & \leq \exp((l-1)\bar{\mu}(I,J) - l\bar{\mu}(I,J) \log l) \\
    & \leq \exp (-\frac{1}{2} l \log l \bar{\mu}(I,J)).
\end{align*}
For a constant $c_5>0$, let
\[
t(I,J) \log t (I,J)=\frac{c_5|J|}{\bar{\mu} (I,J)} \log \frac{n}{|J|},
\]
and let $l(I,J)=\max \{8,t(I,J)\}$. Then, from the previous calculations, we have
\[
 P(e(I,J) \geq l(I,J) \bar{\mu}(I,J)) \leq \exp (-\frac{1}{2} \bar{\mu}(I,J) l(I,J) \log l(I,J) ) \leq c_3|J| \log \frac{n}{|J|}.
\]
From this point onwards identical arguments as those used in \cite{lei2015consistency} can be invoked to complete the proof of Lemma ~\ref{twostatement}.
\end{proof}

\subsubsection*{Proof of Lemma \ref{bounded degree TE}}

\begin{proof}
We use Proposition \ref{prop1}. Let us start with the observation that
\[
E[(d_{E^3})_i]=E[\sum_{j} \sum_{k}E_{ij}E_{jk}E_{ik}] \leq n^2(p^e_{\max})^3 \leq \Delta_{E^3}.
\]
Furthermore, let $I_i$ be the set of all triangles of type $E^3$ incident to a vertex $i$. Let $E^3_{\theta}=E_{ij}E_{jk}E_{ik}$, indexed by $\theta=\{i,j,k\},$ denote a family of indicator random variables. Define a ``good event" $\Gamma$ as before: under the good event, every pair of vertices has at most $C=4\tau_{\max}$ common neighbors. Clearly, two triangles belonging to the set $I_i$ are independent if they do not share any edges. For simplicity, let ``$\sim$" denote a relation such that $\theta_1 \sim \theta_2$ holds if $\theta_1$ and $\theta_2$ share an edge. For any $E^3_{ijk}$, the good event $\Gamma$ restricts the number of triangles of type $E^3$ in the set $I_i$ that are dependent on $E^3_{ijk}$ to $2C$. 

Define $J \subset I$ as
\[
J=\{ \theta : \max_{\theta_1 \in J} |\theta_2 \in J;\theta_1 \text{ and } \theta_2 \text{ share at least one edge } | \leq 2C \}.
\]
Then, we have 
\[
\max_{\theta_2 \in J} \sum_{\theta_1 \in J: \theta_1 \sim \theta_2} E^3_{\theta_1} \leq 2C = 8\tau_{\max},   \quad \mu = E[\sum_{\theta \in I_i} E^3_{\theta}]\Delta_{E^3}.
\]
For $t=\mu=\Delta_{E^3},$ the above results imply
\begin{align*}
    P(\max_{J} \sum_{\theta\in J} E^3_{\theta} \geq 2\Delta_{E^3})  
    & \leq \min \bigg \{\exp \left( -\frac{\Delta_{E^3}^2}{2C(\Delta_{E^3} + \Delta_{E^3}/3)}\right), \left(1+\frac{\Delta_{E^3}}{2\Delta_{E^3}}\right)^{-\Delta_{E^3}/2C} \bigg \}\\
    & = \min \bigg \{\exp \left(- \frac{\Delta_{E^3}}{ \frac{32}{3}\tau_{\max}}\right),2^{-\Delta_{E^3}/8\tau_{\max}} \} \\
    & \leq \exp(-c' \log n) \leq n^{-c'},
\end{align*}
where the last inequality is a consequence of the following argument. If $\tau_{\max}=n(p^e_{\max})^2$, then $\frac{\Delta_{E^3}}{ \tau_{\max}} = np^e_{\max} \geq \log n$ by assumption, and if $\tau_{\max}=\log n$, then $\frac{\Delta_{E^3}}{ \tau_{\max}} \geq \log n$. Under the good event $\Gamma$, we have $I=J^{*}$ and consequently, $\max_{J} \sum_{\alpha \in J} E^3_{\alpha} = (d_{E^3})_i$. 
Then,
\[
P((d_{E^3})_i \geq 2 \Delta_{E^3}) n^{-c'} + P(A \notin \Gamma) \leq n^{-c''},
 \]
since $P(A \notin \Gamma) \leq \exp(-\frac{1}{4} \log n)$.
Taking the union bound over all values of $i$ results in $\max_i (d_{TE})_i \leq c_1 \Delta_{TE}$ with probability at least $1-n^{-c''}$. 
\end{proof}

\subsection*{Proof of Lemma \ref{two statement TE}}

\begin{proof}
We first note that $\bar{\mu}(I,J)=|I||J|n(p^e_{\max})^3$ and $\Delta_{E^3} \geq n^2(p^e_{\max})^3$.

Next, if $|J|>n/e$, then the result of Lemma \ref{bounded degree TE} implies that
\[
   \frac{e(I,J)}{\bar{\mu}(I,J)} =   \frac{e(I,J)}{|I||J|n(p^e_{\max})^3}  \leq \frac{\sum_{i \in I} \max_i (d_{E^3})_i}{n^2p^3_{\max}|I|/e} \leq \frac{|I|c_1\Delta_{E^3}}{n^2p^3_{\max}|I|/e} \leq c_1\,e,
\]
so that in this case (a) holds true.
If $|J| < n/e$, define $S(I,J)$ as the set of all $3$-tuples such that each tuple has one vertex in each of the sets $I$ and $J$.

To prove that the second statement also holds, we cannot invoke the exponential concentration inequality used in the proof of Theorem 1 due to the lack of the independence assumption. Instead, we use Proposition \ref{prop1} on the set $S(I,J)$ of $3$-tuples. 

First, note that 
$$e(I,J) =\sum_{i \in I} \sum_{j \in J} (A_{E^3})_{ij} = \sum_{\theta \in S(I,J)} (A_{E^3})_{\theta}$$ 
and 
$$E(\sum_{\theta \in S} (A_{E^3})_{\theta}) \leq |I||J|n(p^e_{\max})^3 =\bar{\mu}(I,J).$$ 
Define $S^* \subset S$ to be such that any $(A_{E^3})_{\theta}$, for some $\theta\in S^*$, depends only on $\tau_{\max}$ other $(A_{E^3})_{\theta}$ s. We then have $\max_{S^*}\sum_{\theta \in S*(I,J)} (A_{E^3})_{\theta} \leq \tau_{\max}$. 

Next, let $t=(l-1)\bar{\mu}(I,J)$. Then, for some $l\geq 8$,
\begin{align*}
    P(e(I,J) \geq l \bar{\mu}(I,J)) & \leq \exp(-\frac{l\bar{\mu}(I,J) \log l -(l-1)\bar{\mu}(I,J)}{\tau_{\max}}) \\
    & \leq \exp (-\frac{1}{2} \frac{l \log l \bar{\mu}(I,J)}{\tau_{\max}}).
\end{align*}
For a constant $c_3>0$, define $t(I,J)$ according to
\[
t(I,J) \log t (I,J)=\frac{c_3\tau_{\max}|J|}{\bar{\mu} (I,J)} \log \frac{n}{|J|},
\]
and let $l(I,J)=\max \{8,t(I,J)\}$. From the previous calculations, we have 
\[
 P(e(I,J) \geq l(I,J) \bar{\mu}(I,J)) \leq \exp (-\frac{1}{2} \frac{\bar{\mu}(I,J)}{\tau_{\max}} l(I,J) \log l(I,J) ) \leq \exp(-c_3 |J| \log \frac{n}{|J|}).
\]
Following an identical argument as described in~\cite{lei2015consistency}, we have
\[
P(\exists I,J: |I| \leq |J| \leq \frac{n}{e}, \, e(I,J) \geq l(I,J)\bar{\mu}(I,J)) \leq n^{-c_4}.
\]
Therefore, with probability at least $1-n^{-c_4}$, we have $e(I,J) \leq l(I,J)\bar{\mu}(I,J))$. For pairs $\{(I,J): |I| \leq |J| \leq \frac{n}{e}\}$ such that $l(I,J) =8$, we readily have
\[
\frac{e(I,J)}{\bar{\mu}(I,J)}\leq 8.
\]
This establishes that part (a) of the claim is true. For the remaining pairs for which $l(I,J)=t(I,J)$ holds, we have $\frac{e(I,J)}{\bar{\mu}(I,J)} \leq t(I,J)$, and
\[
\frac{e(I,J)}{\bar{\mu}(I,J)} \log \frac{e(I,J)}{\bar{\mu}(I,J)} \leq t(I,J) \log t(I,J) = \frac{c_3\tau_{\max}|J|}{\bar{\mu} (I,J)} \log \frac{n}{|J|},
\]
implying that part (b) of the claim is true as well.
\end{proof}

\section*{Acknowledgement}
The authors would like to thank Prof. Lutz Warnke of Georgia Institute of Technology for explaining how his work may be applied in parts of the analysis. We also appreciate his generous assistance with proving some of the results. The work was supported by the NSF grant CCF 1527636, the Center of Science of Information NSF STC center, and an NIH U01 grant for targeted software development.

\bibliography{triads.bib,Networks.bib,vc.bib,cluster.bib,GraphTheory.bib,triads1.bib}
\end{document}